\documentclass{article}

    \PassOptionsToPackage{numbers, compress,sort}{natbib}

\usepackage[final]{neurips_2025}

\usepackage{wrapfig}

\usepackage{multirow,mathtools }

\usepackage{pifont}
\usepackage{color, colortbl}

\usepackage{blindtext}
\usepackage{lipsum}

\usepackage{multirow}
\usepackage{listings}

\usepackage{bbm}

\usepackage [english]{babel}
\usepackage[autostyle, english = american]{csquotes}

\usepackage{pifont}
\usepackage{url}
\usepackage[most]{tcolorbox}

\usepackage{lipsum}
\usepackage{soul}
\usepackage{xcolor}
\usepackage{wrapfig}
\usepackage{multirow,mathtools } 

\usepackage{adjustbox}
\MakeOuterQuote{"}

\usepackage{microtype}
\usepackage{graphicx}
\usepackage{subfigure}
\usepackage{booktabs} 

\usepackage{hyperref}

\usepackage{amsmath}

\usepackage[capitalize,noabbrev]{cleveref}

\usepackage{nicefrac}       

\usepackage{tablefootnote}

\newcommand{\Def}[0]{\mathrel{\mathop:}=}

\usepackage{pifont}

\usepackage{color, colortbl}
\definecolor{Gray}{gray}{0.93}
\definecolor{Orange}{rgb}{1,0.5,0}
\definecolor{DGray}{gray}{0.83}
\definecolor{LightCyan}{rgb}{0.88,1,1}

\definecolor{WarnREd}{rgb}{1,0.4,0.4}
\definecolor{WarnOrange}{rgb}{1,0.682,0.502}
\definecolor{WarnPink}{rgb}{0.9176, 0.7215, 0.7215}
\definecolor{GoodGreen}{rgb}{0.5019, 0.9215, 0.6039}

\usepackage[T1]{fontenc}

\definecolor{styleblue}{HTML}{504099}
\definecolor{mypurple}{HTML}{9391ff}

\definecolor{bluegray}{rgb}{0.4, 0.6, 0.8}
\definecolor{ceruleanblue}{rgb}{0.16, 0.32, 0.75}

\hypersetup{
colorlinks=true,
citecolor=ceruleanblue,
linkcolor=ceruleanblue,
urlcolor=black
}

\newcommand{\textremarkright}[1]{\textcolor{darkgreen}{\textbf{{#1}}}}
\newcommand{\textremarkwrong}[1]{\textcolor{darkred}{\textbf{{#1}}}}
\newcommand{\textremarkquestion}[1]{\textcolor{darkblue}{\textbf{{#1}}}}
\newcommand{\textremarkrepeat}[1]{\textcolor{darkyellow}{\textbf{{#1}}}}
\definecolor{darkgreen}{rgb}{0.0, 0.45, 0.0}
\definecolor{darkred}{rgb}{0.5, 0.0, 0.0}
\definecolor{darkblue}{rgb}{0.0, 0.0, 0.5}
\definecolor{darkyellow}{rgb}{0.65, 0.65, 0}

\usepackage{amsmath,amsfonts,bm}

\def\eqref#1{(\ref{#1})}

\def\1{\bm{1}}

\newcommand{\Dr}{\mathcal{D_{\mathrm{r}}}}
\newcommand{\Df}{\mathcal{D_{\mathrm{f}}}}

\def\eps{{\epsilon}}

\DeclareMathAlphabet{\mathsfit}{\encodingdefault}{\sfdefault}{m}{sl}
\SetMathAlphabet{\mathsfit}{bold}{\encodingdefault}{\sfdefault}{bx}{n}

\newcommand{\E}{\mathbb{E}}

\newcommand{\R}{\mathbb{R}}

\DeclareMathOperator*{\minimize}{\text{minimize}}

\newcommand{\btheta}{{\boldsymbol{\theta}}}

\newcommand{\ellf}{{\ell_{\mathrm{f}}}}
\newcommand{\ellr}{{\ell_{\mathrm{r}}}}

\newcommand{\widebar}[1]{\overline{#1}}
\newcommand{\bigO}{{\mathcal{\widetilde O}}}
\newcommand{\SL}[1]{\textcolor{orange}{SL: #1}}

\everydisplay{\small}

\title{Simplicity Prevails: Rethinking Negative Preference Optimization for LLM Unlearning}

\author{Chongyu Fan$^{\dag, \star}$ ~~Jiancheng Liu$^{\dag, \star}$ ~~Licong Lin$^{\ddag, \star}$ \\ ~~\textbf{Jinghan Jia}$^{\dag}$ ~~\textbf{Ruiqi Zhang}$^{\ddag}$ ~~\textbf{Song Mei}$^{\ddag}$ ~~\textbf{Sijia Liu}$^{\dag,\S}$\\
  $^\dag$Michigan State University\\
  $^\ddag$University of California, Berkeley\\
  $^\S$IBM Research\\
  $^\star$Equal contributions
}

\begin{document}

\maketitle

\begin{abstract}
This work studies the problem of large language model (LLM) unlearning, aiming to remove unwanted data influences (\textit{e.g.}, copyrighted or harmful content) while preserving model utility. Despite the increasing demand for unlearning, a technically-grounded optimization framework is lacking. Gradient ascent (GA)-type methods, though widely used, are suboptimal as they reverse the learning process without controlling optimization divergence (\textit{i.e.}, deviation from the pre-trained state), leading to risks of model collapse. Negative preference optimization (NPO) has been proposed to address this issue and is considered one of the state-of-the-art LLM unlearning approaches. In this work, we revisit NPO and identify another critical issue: reference model bias. This bias arises from using the reference model (\textit{i.e.}, the model prior to unlearning) to assess unlearning success, which can lead to a misleading impression of the true data-wise unlearning effectiveness.
Specifically, it could cause (a) uneven allocation of optimization power across forget data with varying difficulty levels and (b) ineffective gradient weight smoothing during the early stages of unlearning optimization. To overcome these challenges, we propose a simple yet effective unlearning optimization framework, called SimNPO, showing that  `simplicity' in removing the reliance on a reference model (through the lens of simple preference optimization) benefits unlearning. We provide deeper insights into SimNPO's advantages, including an analysis based on mixtures of Markov chains. Extensive experiments further validate its efficacy on benchmarks like TOFU, MUSE and WMDP. Codes are available at \,\url{https://github.com/OPTML-Group/Unlearn-Simple}.

\end{abstract}




\section{Introduction}
\label{sec: intro}


The rapid advancement of LLMs has raised security and safety concerns, including issues related to copyright violations and sociotechnical harms \citep{sun2024trustllm,wang2023decodingtrust,li2024wmdp,shi2024muse}. However, retraining these models to remove undesirable data influences is often impractical due to the substantial costs and time required for such processes. This gives rise to the problem of \textbf{LLM unlearning} \citep{liu2024rethinking}. 
%
To trace its origins,
the concept of {machine unlearning} was initially developed for data removal to comply with privacy regulations such as the ``right to be forgotten'' \citep{rosen2011right,hoofnagle2019european}, with early studies focusing on vision models \citep{cao2015towards,warnecke2021machine,bourtoule2021machine,thudi2022unrolling,kurmanji2024towards,jia2023model,gandikota2023erasing,fan2023salun}. However, it is soon adapted to LLMs to remove unwanted data and knowledge  \citep{eldan2023whos,yao2023large,liu2024rethinking,shi2024muse,maini2024tofu,zhang2024negative,jia2024soul}.

%

The current optimization foundation for LLM unlearning often relies on \textit{optimization divergence}\footnote{Here, we use ``divergence'' as opposed to ``convergence'' in model training, aiming to reverse learning for unlearning.} from the pre-trained state, which refers to the deviation from the converged pre-trained model to reverse the effects of learning the forgotten data, thereby achieving unlearning
\citep{liu2022continual,maini2024tofu,zhang2024negative}.
%
Nevertheless, the lack of control over the divergence rate in unlearning optimization can lead to either under-forgetting, where insufficient unwanted data influence is removed, or over-forgetting, causing a significant loss of model utility in LLMs. Therefore, \textit{optimization for LLM unlearning is highly non-trivial}. 
%
Negative preference optimization (\textbf{NPO}) \citep{zhang2024negative} emerges as an effective approach for LLM unlearning, as demonstrated by its better control of the divergence rate during unlearning optimization and its
strong performance in current benchmarks such as TOFU \citep{maini2024tofu} and MUSE \citep{shi2024muse}. 
Inspired by direct preference optimization (DPO) \citep{rafailov2024direct}, it treats the forget data points as negative responses, providing a lower-bounded unlearning objective. This also induces a gradient weight smoothing scheme to regulate the speed of divergence.  We refer readers to Sec.\,\ref{sec: background} for details.

Despite the advancements NPO has brought to the optimization foundation for LLM unlearning, our work identifies, for the first time, its potential {limitations} stemming from its reliance on the reference model (\textit{i.e.}, the model prior to unlearning) as the basis for promoting and regulating the optimization divergence. We term this issue \textit{reference model bias}.  See the \textit{conceptual schematic overview} below.

\textbf{Fig.\,\ref{fig: intro_fig}-(a)} illustrates this issue schematically. NPO aims to widen the gap between the unlearned model ($\btheta_{\mathrm{NPO}}$) and the reference model ($\btheta_{\mathrm{ref}}$). However, the prediction confidence of $\btheta_{\mathrm{ref}}$ varies across samples, as illustrated by the ``hard'' vs. ``easy''  unlearning examples along the green line in Fig.\,\ref{fig: intro_fig}-(a). Specifically, ``hard'' examples are those whose predictions under $\btheta_{\mathrm{ref}}$ lie far from the unlearning decision boundary, making them more difficult to forget. In contrast, ``easy'' examples are already close to the boundary, where further increasing the gap between the unlearned model and $\btheta_{\mathrm{ref}}$ could become unnecessary.
Yet, NPO may blindly increase the deviation from $\btheta_{\mathrm{ref}}$ (as shown by the blue line in Fig.\,\ref{fig: intro_fig}-(a)), causing ``easy'' examples to move unnecessarily far beyond the unlearning boundary. Meanwhile,``hard'' examples remain far from the boundary and require more targeted effort to forget. 
In other words, relying on the reference model can result in suboptimal unlearning power allocation due to its uniform, deviation-based strategy.


\begin{figure*}[t]
\vspace*{-3mm}
\centering
\begin{tabular}{cccc}
\includegraphics[width=0.18\textwidth,height=!]{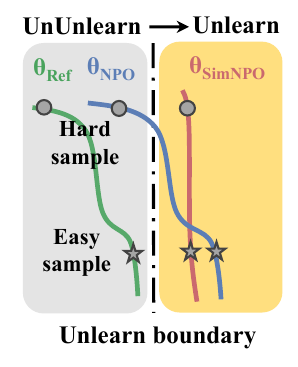} 
&
\hspace*{-3mm}
\includegraphics[width=0.25\textwidth,height=!]{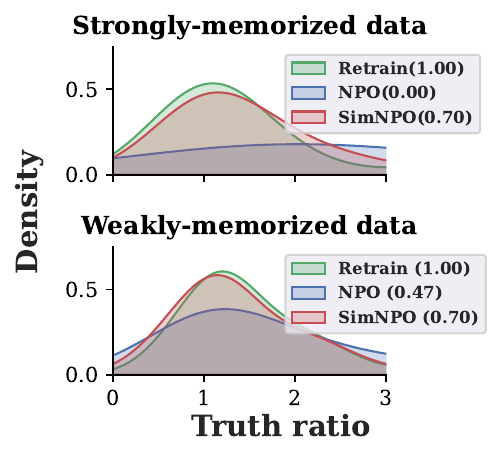} 
&
\hspace*{-3mm}
\includegraphics[width=0.265\textwidth,height=!]{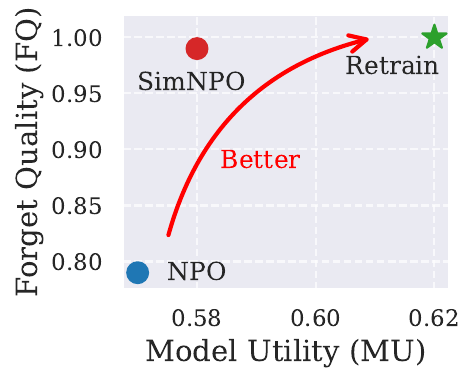}
&
\hspace*{-6mm}
\includegraphics[width=0.245\textwidth,height=!]{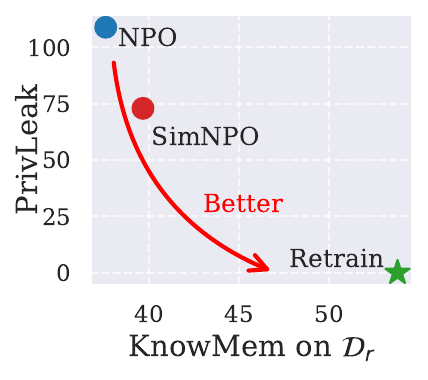}
\vspace*{-1mm}\\
\scriptsize{(a) Schematic overview} & \scriptsize{(b) FQ vs. data memorization} & 
\scriptsize{(c) TOFU, LLaMA-2-chat 7B} & \scriptsize{(d) MUSE, LLaMA-2 7B }
\end{tabular}
\vspace*{-3mm}
\caption{\small{
{\textit{(a)} {Systematic overview} of an LLM ($\btheta$) post-unlearning using the proposed SimNPO, compared to NPO \citep{zhang2024negative} and the reference model. \textit{(b)} Truth ratio distribution of strongly-memorized forget data points and weakly-memorized data for NPO, SimNPO, and Retrain on the TOFU Forget05 dataset \citep{maini2024tofu} under LLaMA-2-chat 7B; See Sec.\,\ref{sec: NPO_limitations} for more details. As shown, SimNPO achieves better forget quality (FQ, the number after method) than NPO and exhibits a truth ratio distribution closer to Retrain. Note that FQ is a statistical measure quantifying the closeness between the truth ratio distribution of an unlearned model and that of Retrain (with FQ$=1$ representing optimal unlearning).
\textit{(c) \& (d)} {Experiment highlights} on  TOFU Forget05 and MUSE News datasets \citep{shi2024muse}. Unlearning effectiveness is measured by FQ for TOFU and PrivLeak for MUSE, while utility preservation is evaluated using model utility for TOFU and KnowMem on retain data for MUSE (see Table\,\ref{tab:tasks_evaluations}). In both tasks, Retrain is the gold standard for unlearning.
}}}
\label{fig: intro_fig}
\vspace*{-3mm}
\end{figure*}

Throughout this work, we ask:
\begin{tcolorbox}[before skip=2mm, after skip=0.0cm, boxsep=0.0cm, middle=0.0cm, top=0.05cm, bottom=0.05cm, boxrule=0.6pt]
\begin{center}
     \textit{\textbf{(Q)} How can we identify and address the limitations of NPO to enhance its effectiveness?}
\end{center}
\end{tcolorbox} 
\vspace*{2mm}

In response to \textbf{(Q)}, we propose a simple yet effective unlearning optimization framework, termed \textbf{SimNPO}, demonstrating that {properly} removing reliance on a reference model can significantly enhance unlearning. This approach also draws inspiration from simple preference optimization in LLM alignment \citep{meng2024simpo}. Additionally, we will provide detailed  insights into how SimNPO overcomes the limitations of NPO caused by reference model bias. As shown schematically in {Fig.\,\ref{fig: intro_fig}-(a)}, SimNPO outperforms NPO by more accurately identifying the difficulty of unlearning data (\textit{i.e.}, hard vs. easy samples) and allocating optimization power more effectively across different forget samples.
\textbf{Fig.\,\ref{fig: intro_fig}-(b)} provides experimental evidence, which will be provided in \textbf{Sec.\,\ref{sec: NPO_limitations}},  by comparing the unlearning performance of NPO and SimNPO across forget data points with their unlearning difficulty levels indicated by their \textit{memorization levels}.
The rationale is that the reference model demonstrates varying levels of memorization across different forget samples, making \textit{strongly-memorized} samples \textit{harder} to unlearn and \textit{weakly-memorized} samples \textit{easier} to unlearn. However, NPO may blindly over-allocate unlearning power to these easier samples, thereby hindering the effective unlearning of harder ones. This explains why Fig.\,\ref{fig: intro_fig}-(b) shows that NPO performs worse than SimNPO in the strongly-memorized (hard) forget data, as evidenced by a greater deviation from \textbf{Retrain}.

In summary, ours contributions are outlined below:

$\bullet$  We revisit the NPO framework and identify its potential weakness--reference model bias--in LLM unlearning, which 
can lead to issues such as sensitivity to the reference model's response quality and ineffective gradient weight smoothing. 
We reveal and justify this bias through a series of analyses/examples, including reference model perturbation, the relationship between unlearning and data memorization, and the impact of forget data length on unlearning.

$\bullet$ Building on insights into NPO's limitations, we propose an improved LLM unlearning approach, SimNPO, which extends NPO using a reference-free optimization framework, simple preference optimization \citep{meng2024simpo}. Despite its simplicity, our methodology is grounded in a rigorous technical rationale, as supported by additional synthetic studies and theoretical insights.



$\bullet$ We conduct extensive experiments to demonstrate the improvements of SimNPO over NPO across various scenarios, including  TOFU \citep{maini2024tofu}, MUSE \citep{shi2024muse},  WMDP \citep{li2024wmdp}, and defending against relearning-based attacks \citep{lynch2024eight,hu2024jogging}.
Some experiment highlights on TOFU and MUSE unlearning benchmark datasets are showcased in
\textbf{Fig.\,\ref{fig: intro_fig}-(c,d)}.






\section{Related work}

\noindent 
\textbf{Machine unlearning.} 
From the perspective of whether the forget data can be inferred from the unlearned model in terms of membership (\textit{i.e.}, a data privacy viewpoint), 
the widely adopted gold standard for machine unlearning is `Retrain' \cite{cao2015towards,thudi2022unrolling,jia2023model}, which we also adopt in this work. Also known as \textit{exact} unlearning, this approach retrains the model from scratch on the original training set with the forget data excluded.
%
However, exact unlearning is challenging in practice due to the assumption for access to the full training set  
and the high computational cost of retraining. To address these challenges, various \textit{approximate} unlearning methods have been developed \citep{nguyen2022survey,bourtoule2021machine,triantafillou2024we}. These approaches typically involve model fine-tuning or editing, applied to the pre-trained model, based on the unlearning request. Their effectiveness has been shown in different application domains, including image classification \citep{liu2022backdoor,jia2023model,kurmanji2024towards,fan2024challenging}, image generation \citep{gandikota2023erasing,fan2023salun,zhang2024unlearncanvas}, federated learning \citep{liu2022right,halimi2022federated,jin2023forgettable}, and graph neural networks \citep{chen2022graph,chien2022certified,wu2023certified}. 


\noindent
\textbf{LLM unlearning.}
There has also been a growing body of research focusing on LLM unlearning \citep{lu2022quark,jang2022knowledge,kumar2022privacy,zhang2023forget,pawelczyk2023context,eldan2023whos,ishibashi2023knowledge,yao2023large,maini2024tofu,zhang2024negative,li2024wmdp,wang2024large,jia2024soul,liu2024rethinking,liu2024large,thaker2024guardrail,kadhe2024split,gu2024meow,wang2024rkld,mekala2024alternate}, aiming to effectively remove undesired data influences and/or model behaviors while preserving the  utility for unrelated knowledge generation, and maintaining efficiency without the need for retraining. 
Applications of unlearning in LLMs are diverse, from safeguarding copyrighted and personally identifiable information \citep{jang2022knowledge,eldan2023whos,wu2023depn}, to preventing LLMs from creating cyberattacks or bioweapons \citep{barrett2023identifying,li2024wmdp}, and reducing the production of offensive, biased, or misleading content \citep{lu2022quark,yu2023unlearning,yao2023large}. 
Current unlearning approaches include model optimization-based methods \citep{ilharco2022editing,liu2022continual,yao2023large,eldan2023whos,jia2024soul,zhang2024negative,li2024wmdp,gu2024meow,wang2024rkld,mekala2024alternate} and input prompt or in-context learning-based techniques \citep{thaker2024guardrail,pawelczyk2023context,liu2024large}.
However, many lack effectiveness, leading to either under-forgetting or over-forgetting, as shown by recent LLM unlearning benchmarks such as TOFU for fictitious unlearning \citep{maini2024tofu} and MUSE for private or copyrighted information removal \citep{shi2024muse}.
Recent studies also show that even after unlearning, models can remain vulnerable to adversarial attacks \citep{schwarzschild2024rethinking,patil2023can,lynch2024eight} or relearning from a small number of forget data \citep{hu2024jogging,lynch2024eight}.
This evidence suggests that effective unlearning for LLMs is far from trivial. Among current efforts, NPO (negative preference optimization) \citep{zhang2024negative}  stands out as a promising method. However, we will show that the advantages of NPO can be limited by the presence of reference model bias (Sec.\,\ref{sec: NPO_limitations}).
\noindent
\textbf{Preference optimization.}
In this work, we advance LLM unlearning through the lens of preference optimization. This is motivated by aligning LLMs with human values, known as reinforcement learning from human feedback (RLHF) \citep{christiano2017deep, ziegler2019fine, ouyang2022training}. However, online preference optimization algorithms are often complex and challenging to optimize \citep{santacroce2023efficient, zheng2023secrets}, driving interest in more efficient offline alternatives. Direct preference optimization (\textbf{DPO}) \citep{rafailov2024direct} introduced an offline approach that eliminates the need for a reward model, sparking the development of several reward-free offline preference objectives \citep{zhao2023slic, azar2024general, hong2024reference, ethayarajh2024kto, meng2024simpo, yuan2024rrhf}. Notable methods include RRHF \citep{yuan2024rrhf}, SLic-HF \citep{zhao2023slic}, IPO \citep{azar2024general}, KTO \citep{ethayarajh2024kto},   ORPO \citep{hong2024reference}, and  SimPO \citep{meng2024simpo}. Among these methods, SimPO is a reference-free, length-normalized variant of DPO, and we will demonstrate that it is well-suited
for integrating into LLM unlearning and improving NPO.

\vspace*{-1mm}
\section{A Primer on LLM Unlearning}
\label{sec: background}
\vspace*{-1mm}

\noindent
\textbf{Problem formulation.}
%
%
Unlearning tasks can take various forms and are typically associated with a specific set of data points to be removed, known as the \textit{forget set} ($\Df$). These tasks often require a complementary set of non-forgotten data points, known as the \textit{retain set} ($\Dr$), to preserve model utility by penalizing the divergence caused by unlearning.
As a result,  the problem of LLM unlearning can be cast as a regularized optimization problem that balances  the forget and retain objectives \citep{liu2024rethinking,zhang2024negative}:
 \begin{align}
\hspace*{-3mm}
\begin{array}{l}
 \displaystyle \minimize_{\boldsymbol{\theta}}    \,\, 
 \mathbb E_{(x, y) \in \Df} [ \ellf(y | x; \boldsymbol{\theta}) ]
 + \lambda 
 \mathbb E_{(x, y) \in \Dr} [ \ellr( y | x; \boldsymbol{\theta}) ] 
 ,
\end{array}
\hspace*{-3mm}
\label{eq: prob_LLM_MU}
\end{align}
where $\btheta$ represents the model parameters to be updated during unlearning,  $\lambda \geq 0$ is a regularization parameter to penalize the `divergence'
of unlearning, and $\ellf$ and $\ellr$ represent forget and retain losses incurred when using model parameters $\boldsymbol{\theta}$ to generate $y$ given the input $x$.

Substantial research has focused on designing and analyzing appropriate forget and retain loss functions to solve  problem \eqref{eq: prob_LLM_MU} \citep{liu2024rethinking,yao2023large,zhang2024negative,maini2024tofu,shi2024muse,eldan2023whos,jia2024soul}.
For instance, let $\pi_\mathrm{\btheta}(y | x)$ represent the prediction probability of the model $\btheta$ given the input-response pair $(x, y)$.
The retain loss is typically chosen as the cross-entropy-based sequence  prediction loss, 
$\ellr(y | x, \btheta) = -\log \pi_\mathrm{\btheta}(y | x)$, whose minimization encourages the model 
to perform well on the retain data $(x, y) \in \Dr$. In \eqref{eq: prob_LLM_MU},  if we specify the forget loss as the \textit{negative} token prediction loss $\ellf(y | x, \btheta) =  \log \pi_\mathrm{\btheta}(y | x)$, whose minimization then \textit{discourages} the model from learning the forget data $(x, y) \in \Df$. Minimizing such a forget loss is known as the \textit{gradient ascent} (\textbf{GA}) method \citep{maini2024tofu,thudi2022unrolling}. Similarly, minimizing the regularized loss that integrates GA with the  
 retain loss is known as the \textit{gradient difference} (\textbf{GradDiff}) method \citep{liu2022continual,maini2024tofu,yao2023large}.

\noindent
\textbf{Negative preference optimization (NPO).}
A popular optimization framework for solving problem \eqref{eq: prob_LLM_MU} is NPO \citep{zhang2024negative}.
It treats the forget data as negative examples in DPO \citep{rafailov2024direct}, 
transforming the {unbounded} 
GA-based forget loss into a \ding{172} \textit{bounded loss from below},
which helps prevent catastrophic collapse, and an  \ding{173} \textit{adaptive weight smoothing} applied to the forget loss gradients, enabling more controlled divergence speed in unlearning. 

These benefits can be clearly seen from the NPO loss and its gradient as follows: 
\begin{align}
  & \hspace*{-2mm} \ell_{\mathrm{NPO}}(\boldsymbol{\theta}) =  \mathbb E_{(x,y) \in \Df} \underbrace{\left [    - \frac{2}{\beta} \log \sigma  \left ( - \beta \log \left ( \frac{\pi_{\btheta} (y | x) }{\pi_{\mathrm{ref}} (y | x)}\right ) \right ) 
    \right ]}_\text{\ding{172} $\Def \ellf(y|x; \btheta)$, the specified forget loss in \eqref{eq: prob_LLM_MU} 
}  \hspace*{-3mm} \label{eq:NPO_loss} \\
&   \hspace*{-2mm} \nabla_{\btheta} \ell_{\mathrm{NPO}}(\boldsymbol{\theta}) = 
 \mathbb E_{(x,y) \in \Df}  \left[  \underbrace{
\left ( \frac{2
\pi_{\btheta}(y | x )^\beta
}
{ \pi_{\btheta}(y | x )^\beta + \pi_{\mathrm{ref}}(y | x )^\beta }  \right) }_\text{\ding{173} $\Def w_{\btheta}(x, y)$, adaptive weight}
\cdot
\underbrace{\nabla_\btheta \log\pi_{\btheta}(y | x)}_\text{GA}
\right ]
\hspace*{-3mm}
\label{eq:NPO_grads}
\end{align}
where $\sigma (t) = 1/(1+e^{-t})$ is the sigmoid function, $\beta > 0 $ is the temperature parameter  and $\pi_{\mathrm{ref}}$ is the \textbf{reference model} given by the initial model prior to unlearning. 
Additional insights into \ding{172}-\ding{173} are given below.

\ding{172}
 From \eqref{eq:NPO_loss}, the NPO-type forget loss is bounded below by $0$, \textit{i.e.}, $\ellf(y|x; \btheta) \geq 0$, whereas the GA-type forget loss, $\ellf(y|x, \btheta) = \log \pi_\mathrm{\btheta}(y | x)$,  has no lower bound. 
Moreover, minimizing it towards $\ellf(y|x; \btheta) \to 0$ drives the prediction probability $\pi_{\btheta}(y|x)$ to decrease, widening the gap between the prediction probability and the reference model on the forget set, \textit{i.e.}, $\pi_{\btheta}(y|x) \ll \pi_{\mathrm{ref}}(y|x)$.


\ding{173} As seen in \eqref{eq:NPO_grads}, the adaptive weight $w_{\btheta}(x,y)$ is typically less than $1$ since $\pi_{\btheta}(y|x) < \pi_{\mathrm{ref}}(y|x)$ for forgetting. Consequently, NPO's gradient yields a more controlled and gradual divergence speed (\textit{i.e.}, deviation from the reference model), compared to GA (with $w_{\btheta}(x,y) = 1$).

In this paper, NPO will serve as the primary baseline for LLM unlearning. Its implementation follows the {regularized optimization} in \eqref{eq: prob_LLM_MU}, where the forget loss $\ellf$ is defined as in \eqref{eq:NPO_loss} and  the retain loss $\ellr$ is the token prediction loss  $\ellr(y|x, \btheta) = - \log \pi_\mathrm{\btheta}(y | x)$ applied to the retain set.

\noindent
\textbf{LLM unlearning tasks and evaluations.}
Given that the assessment of LLM unlearning may rely on specific tasks, we next introduce the unlearning tasks and evaluation metrics that this work covers. (1) \textbf{TOFU} \citep{maini2024tofu} considers fictitious unlearning on a synthetic Q\&A dataset. (2) \textbf{MUSE} \citep{shi2024muse} is designed to remove verbatim or knowledge memorization from News and Books datasets, including both verbatim texts and knowledge sets for unlearning evaluation.   (3) \textbf{WMDP} \citep{li2024wmdp} aims to prevent LLMs from generating hazardous content in domains such as biology, cybersecurity, and chemistry. 
%
%
Despite the differences in evaluation metrics across the above tasks, the assessment broadly falls into two categories. (1) \textbf{Unlearning effectiveness}  measures how faithfully undesired data influences or model capabilities are removed. For example, it is assessed by the \textit{forget quality} (FQ) metric in TOFU, which uses a $p$-value to test the indistinguishability between the post-unlearning model and a model retrained on the retain set only, and by \textit{privacy leakage} (PrivLeak) in MUSE, which measures the likelihood of detecting that the model was ever
trained on the forget set. (2) \textbf{Utility preservation}  evaluates the post-unlearning performance on standard utility tasks.  See 
\textbf{Table\,\ref{tab:tasks_evaluations}} in \textbf{Appendix\,\ref{app: evaluation}} for a summary of the unlearning tasks and evaluation metrics.

\section{Uncovering Reference Model Bias in NPO}
\label{sec: NPO_limitations}
\vspace*{-2mm}
In this section, we highlight a key weakness of NPO, which we term `\textit{reference model bias}': The incorporation of the reference model  in NPO biases the unlearning objective towards enlarging the distance relative to this reference model.
As noted in \eqref{eq:NPO_loss}, minimizing the NPO loss drives $\pi_{\btheta}(y|x) \ll \pi_{\mathrm{ref}}(y|x)$. However, using $\pi_{\mathrm{ref}}$ as the basis for NPO's unlearning criterion can introduce negative effects (L1)--(L2), which we will detail later. 

Before that, we present a warm-up study to illustrate NPO’s sensitivity to the choice of the reference model ($\btheta_{\mathrm{ref}}$, used interchangeably with $\pi_{\mathrm{ref}}$). Specifically, we construct a perturbed reference model, $\btheta_{\mathrm{ref}}^\prime$, by averaging the original reference model $\btheta_{\mathrm{ref}}$ with a randomly weighted model, whose weights are drawn from a standard Gaussian distribution with zero mean and variance. We then apply NPO using $\btheta_{\mathrm{ref}}^\prime$ as the reference on the TOFU Forget05 dataset, following the same setup as in Fig.\,\ref{fig: intro_fig}-(c). We find that there exists a substantial drop in forget quality--from 0.79 (with $\btheta_{\mathrm{ref}}$) to 0.27 (with $\btheta_{\mathrm{ref}}^\prime$), while the model utility remains nearly unchanged (0.52 w/ $\btheta_{\mathrm{ref}}^\prime$ vs. 0.57 w/ $\btheta_{\mathrm{ref}}$). 
We refer readers to \textbf{Fig.\,\ref{fig: npo_reference_model}} in \textbf{Appendix\,\ref{app: npo_reference_model}} for the detailed comparison. 
This preliminary study highlights the critical influence of the reference model on NPO’s unlearning effectiveness. Thus, a deeper investigation into the use of the reference model could offer valuable insights for improving the unlearning optimization framework.

Next, we elaborate on the limitations (L1)--(L2) introduced by the reference model in  NPO.

\noindent
\textbf{(L1) 
Challenge of uneven allocation of unlearning power across forget data. 
}
At first glance, driving the unlearned model to deviate from the reference model in NPO, \textit{i.e.}, promoting $\pi_{\btheta}(y|x) \ll \pi_{\mathrm{ref}}(y|x)$, seems desirable for unlearning on the forget set.
%
However, the over-reliance on $\pi_{\mathrm{ref}}$ can overshadow the true sample-specific unlearning difficulty, leading to an uneven allocation of unlearning power.
We elaborate on this issue through two examples. 

\begin{wrapfigure}{r}{.25\textwidth}
\vspace*{-9mm}
\centerline{
\hspace*{0mm}\includegraphics[width=.25\textwidth,height=!]{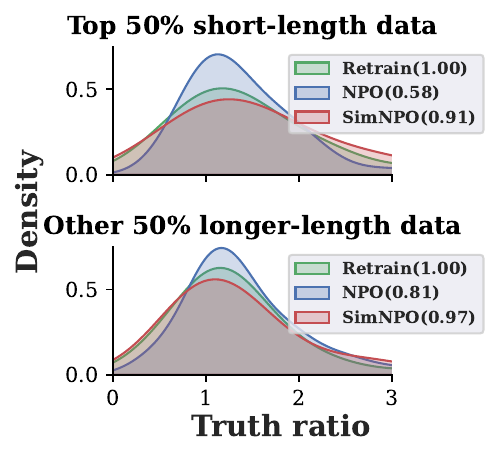}
}
\vspace*{-2mm}
\caption{
\small{Truth ratio distribution of short/long forget data for NPO, SimNPO, and Retrain on TOFU Forget05. The figure format follows Fig.\,\ref{fig: intro_fig}-(b).}}
\label{fig: dist_short_long}
\vspace*{-5mm}
\end{wrapfigure}

\textit{(Example 1: Unlearning strongly vs. weakly-memorized forget data.)}
We first explain (L1) from the perspective of unlearning vs. data memorization. Consider two forget sets, ${\mathcal{D}_{\mathrm{f},1}}$ and ${\mathcal{D}_{\mathrm{f},2}}$, where ${\mathcal{D}_{\mathrm{f},1}}$ is more strongly memorized by the model than ${\mathcal{D}_{\mathrm{f},2}}$. 
To support these memorization levels, we provide detailed experimental settings in \textbf{Appendix\,\ref{app: strongly_weakly_memorized}}.
With this setup, the prediction loss on ${\mathcal{D}_{\mathrm{f},1}}$ is smaller, leading to a higher prediction probability $\pi_{\mathrm{ref}}$. Accordingly, the NPO gradient smoothing term in \eqref{eq:NPO_grads} becomes smaller for ${\mathcal{D}_{\mathrm{f},1}}$, meaning NPO allocates less first-order optimization power to it. However, ${\mathcal{D}_{\mathrm{f},1}}$, being strongly memorized, should ideally receive more unlearning power. As a result, this uneven focus hinders NPO's ability to effectively forget $\mathcal{D}_{\mathrm{f},1}$, potentially causing under-unlearning and reducing the FQ of $\mathcal{D}_{\mathrm{f},1}$ to nearly zero. 
See  \textbf{Fig.\,\ref{fig: intro_fig}-(b)} and \textbf{Table\,\ref{tab: strongly_weakly_memorized}}  for results.

\textit{{(Example 2: Unlearning short vs. long-response data.)}} In this example, we evaluate unlearning performance across different types of forget data, categorized by their response lengths (\textit{i.e.}, short vs. long). The motivation stems from the observation that the reference model may exhibit a bias toward generating longer, yet lower-quality, responses \cite{meng2024simpo}.
\textbf{Fig.\,\ref{fig: dist_short_long}} shows that NPO exhibits a greater distance from Retrain when unlearning the top 50\% shortest-length forget data, resulting in a lower FQ (forget quality) of $0.58$. In contrast, NPO performs better unlearning for the longer 50\% of the forget set, yielding a higher FQ of $0.81$. The ineffectiveness of NPO in unlearning forget data with short responses will be further analyzed through the lens of a mixture of Markov chains in Sec.\,\ref{sec: SimNPO}.

\begin{figure*}[htb]
\centering
\includegraphics[width=0.35\textwidth]{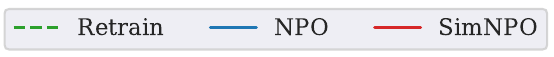}\\
\vspace{-1mm}
\begin{tabular}{cccc}
\hspace*{-3mm}\includegraphics[width=0.21\textwidth]{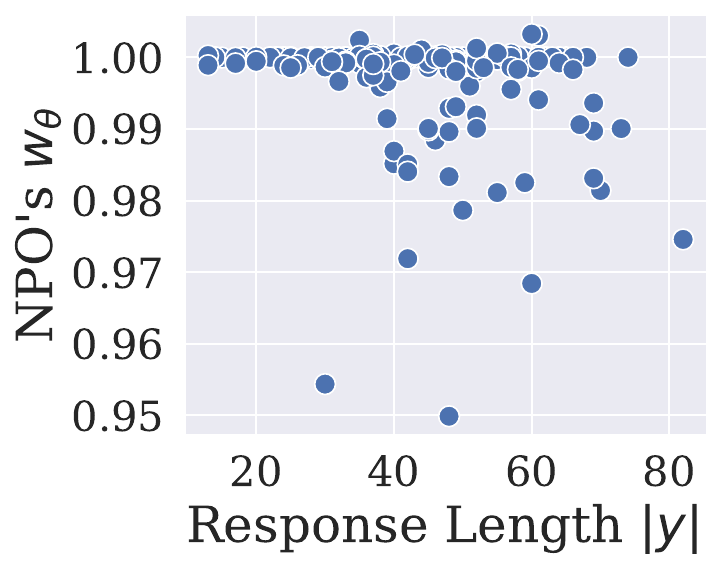} &
\hspace*{-3.9mm}
\includegraphics[width=0.29\textwidth]{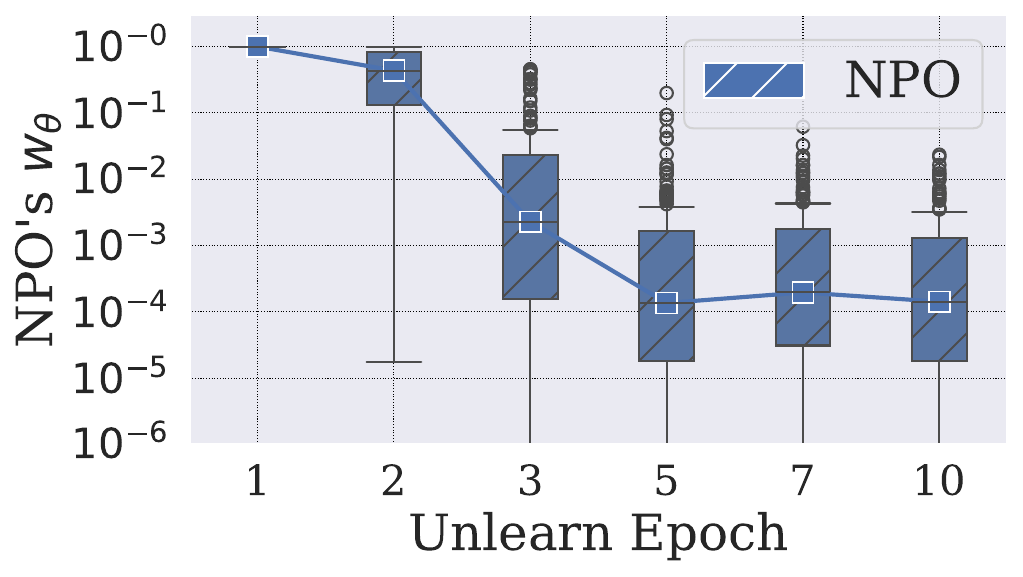} &
\hspace*{-3.9mm}
\includegraphics[width=0.22\textwidth]{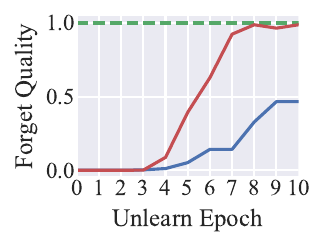} &
\hspace*{-5mm}
\includegraphics[width=0.22\textwidth]{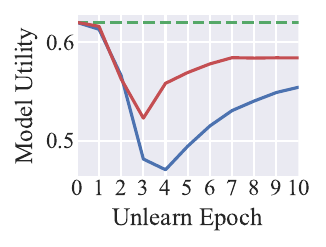}
\vspace*{-1mm}\\
\scriptsize{(a) ${w}_\btheta$ of NPO at epoch 1} & \scriptsize{(b) Trajectory of ${w}_\btheta$ vs. epochs} &
 \scriptsize{ (c) Forget quality vs. epochs} &
 \scriptsize{ (d) Model utility vs. epochs}
\end{tabular}
\vspace*{-2.5mm}
\caption{
\small{
Experimental evidence of ineffective weight smoothing and utility-drop for NPO on TOFU Forget05
(a) NPO's gradient weights ($w_\btheta$) at epoch 1 vs. response length $|y|$.
(b) Trajectory of $w_\btheta$ for NPO over unlearning epochs, where box plot represents the distribution of gradient weights over forget samples.
(c)-(d) Forget quality and model utility of NPO vs. epochs.
}
}
\label{fig: NPO_grad_weights}
\end{figure*}

\noindent
\textbf{(L2) Lack of gradient weight smoothing in the early stages of unlearning.} 
Another issue introduced by the reference model $\pi_{\mathrm{ref}}$ concerns {the effectiveness of NPO's gradient weight smoothing}, \textit{i.e.}, $w_{\btheta}(x,y) = ({2
\pi_{\btheta}(y | x )^\beta
})/
({ \pi_{\btheta}(y | x )^\beta + \pi_{\mathrm{ref}}(y | x )^\beta })
$ in \eqref{eq:NPO_grads}. During the {early optimization stage} of NPO, we find $w_{\btheta}(x,y)   \approx 1$ regardless of the varying data-specific unlearning difficulties since the initialization of the unlearned model $\btheta$ is given by the reference model. \textbf{Fig.\,\ref{fig: NPO_grad_weights}-(a,b)} support this finding by displaying the gradient smoothing weights of NPO at epoch one for forget data with varying response lengths (Fig.\,\ref{fig: NPO_grad_weights}a), as analyzed in Example 2, and their trajectory over the course of unlearning epochs (Fig.\,\ref{fig: NPO_grad_weights}b). 
As shown, the gradient smoothing weights of NPO show large variance, but most values are concentrated around $w_{\btheta}(x,y) \approx 1$ at epoch one.
This implies that NPO behaves similarly to GA in the early stage of unlearning, potentially causing a large utility drop even if the weight decreases in later optimization. 
\textbf{Fig.\,\ref{fig: NPO_grad_weights}-(c,d)} justify the above by presenting FQ and model utility of NPO on TOFU against unlearning epochs. As shown, NPO tends to cause a larger utility drop at early epochs compared to \textit{SimNPO}, the improved alternative to NPO in Sec.\,\ref{sec: SimNPO}. 


%


\vspace*{-1mm}
\section{SimNPO: Method and Rationale
}
\vspace*{-1mm}
\label{sec: SimNPO}

\noindent
\textbf{Motivation of SimNPO and its forget objective.}
The simplest solution to mitigating NPO's reference model bias is to directly remove $\pi_\mathrm{ref}$ from the gradient in   \eqref{eq:NPO_grads}, setting $\pi_\mathrm{ref} = 0$. However, this variant would be \textit{ineffective}, as  the reference-free gradient reduces to GA, with $w_{\btheta}(x,y) = 1$. 
This negates NPO's advantages. 
To develop a better solution for improving NPO, we revisit the context of preference optimization and investigate whether the reference model can be excluded while still retaining the unlearning benefits provided by NPO. Our idea parallels how NPO was originally inspired by DPO \citep{rafailov2024direct}. 
We adopt SimPO \citep{meng2024simpo}, a reference-free alternative to DPO, as the optimization framework for unlearning, leading to the  \textbf{SimNPO} (simple NPO) method.

The \textit{key difference} between SimPO and DPO lies in their reward formulation for preference optimization. In DPO, the  reward formulation is given by the comparison with the reference model, \textit{i.e.}, $\beta \log ( {\pi_{\btheta}(y | x)}/{\pi_{\mathrm{ref}}(y | x)} )$. This formulation was  used by NPO. In contrast, 
SimPO   takes a \textit{reference-free but length-normalized} reward formulation:
$(\beta/|y|) \log  {\pi_{\btheta}(y | x)}$, where $|y|$ denotes the response length.  


Taking the inspiration of SimPO, we can mitigate the reference model bias in NPO by replacing its reward formulation $\beta \log ( {\pi_{\btheta}(y | x)}/{\pi_{\mathrm{ref}}(y | x)} )$ in \eqref{eq:NPO_loss} with the SimPO-based reward formulation $(\beta/|y|)  \log ( {\pi_{\btheta}(y | x)}) $. This modification transforms  \eqref{eq:NPO_loss}  into the \textbf{SimNPO loss}:
 \begin{align}
 \ell_{\mathrm{SimNPO}}(\boldsymbol{\theta}) \hspace*{-1mm}= \hspace*{-1mm} \mathbb E_{(x,y) \in \Df} \hspace*{-1mm} \left [    - \frac{2}{\beta} \log \sigma  \left ( - \frac{\beta}{|y|}\log  \pi_{\btheta} (y | x) - \gamma  \right ) 
    \right ]  \label{eq:SimNPO_loss} 
\end{align}
where  $\gamma\geq0$ is the reward margin parameter, inherited from SimPO, which defines the margin of preference for a desired response over a dispreferred one. However, unless otherwise specified, we set $\gamma = 0$ to align with the NPO loss \eqref{eq:NPO_loss}. This is also desired because $\gamma$ introduces a margin to the prediction loss $-(\beta/|y|)\log \pi_{\btheta}(y|x)$. Consequently, a larger $\gamma$ requires greater compensation to further suppress token prediction, enforcing a stricter unlearning condition. This can accelerate the utility drop during unlearning. {See \textbf{Fig.\,\ref{fig: hyper_parameter_SimNPO} of Appendix\,\ref{app: ablation_hyper}} for the ablation study of hyperparameters.} The SimNPO loss \eqref{eq:SimNPO_loss}, when integrated  in \eqref{eq: prob_LLM_MU}, forms the SimNPO method.

\noindent
\textbf{Insights into SimNPO: Addressing NPO's limitations one by one.}
Similar to  NPO, the SimNPO loss \eqref{eq:SimNPO_loss} is {bounded from below}, with a minimum value of $0$. Approaching this minimum drives the unlearning. 
%
However,
the \textit{key distinction} of SimNPO from NPO is its forget data-aware, length-normalized reward formulation, $(\beta/|y|) \log {\pi_{\btheta}(y | x)}$  in \eqref{eq:SimNPO_loss}. This results in an improved gradient smoothing scheme. 
Specifically, the gradient of the SimNPO loss (with $\gamma = 0$) yields: 
%
\begin{align}
&\nabla_{\btheta}\ell_{\mathrm{SimNPO}}(\btheta)
= \mathbb E_{(x,y) \in \Df}  {\LARGE [} 
\underbrace{
\frac{
2 (\pi_{\btheta}(y|x))^{\beta/|y|} 
}{1+ (\pi_{\btheta}(y|x))^{\beta/|y|}   }
\cdot
\frac{1}{|y|}
}_\text{$\Def w_{\btheta}^\prime(x, y)$}
\cdot
\nabla_\btheta\log\pi_{\btheta}(y|x)
{ \LARGE ] }. \label{eq: grad_SimNPO_Nogamma}
\end{align} 
See {Appendix\,\ref{app: grad_deri}} for derivation. 
Similar to NPO in \eqref{eq:NPO_grads}, the gradient in \eqref{eq: grad_SimNPO_Nogamma} can be divided into two components: weight smoothing ($w_{\btheta}^\prime$) and GA. 
However, in SimNPO, the weight smoothing is \textit{no longer influenced by the reference model and is instead normalized by the length $|y|$}. This introduces two key advantages (a)-(b) below, in response to NPO's limitations (L1)-(L2).

(a)  SimNPO leverages the (data-specific) response length as a guide for unlearning power allocation. For instance, when $|y|$ is large, less optimization power is allocated, helping to avoid the uneven unlearning power allocation across forget data with varying response lengths, as exemplified in Fig.\,\ref{fig: dist_short_long}.
 In the extreme case where $\beta \to 0$, the SimNPO's gradient reduces to a \textit{weighted GA}: $\nabla_{\btheta}\ell_{\mathrm{SimNPO}}(\btheta) \to  \mathbb E_{(x,y) \in \Df} [1/|y| \nabla_\btheta\log\pi_{\btheta}(y|x)]$. This is different from NPO, which becomes GA as $\beta \to 0$. 
{\textbf{Fig.\,\ref{fig: illustration_length_normalization}}} in \textbf{Appendix\,\ref{app: length_normalization}} empirically demonstrates the advantage of length normalization in SimNPO for unlearning.
%
As shown, SimNPO outperforms NPO in both forget quality and model utility, coming closest to Retrain. Even in the special case where $\beta = 0$ (\textit{i.e.}, Weighted-GradDiff), the length normalization provides benefits over GradDiff.

\begin{figure}[htb]
\centering
\hspace*{-5mm}
\includegraphics[width=0.8\textwidth]{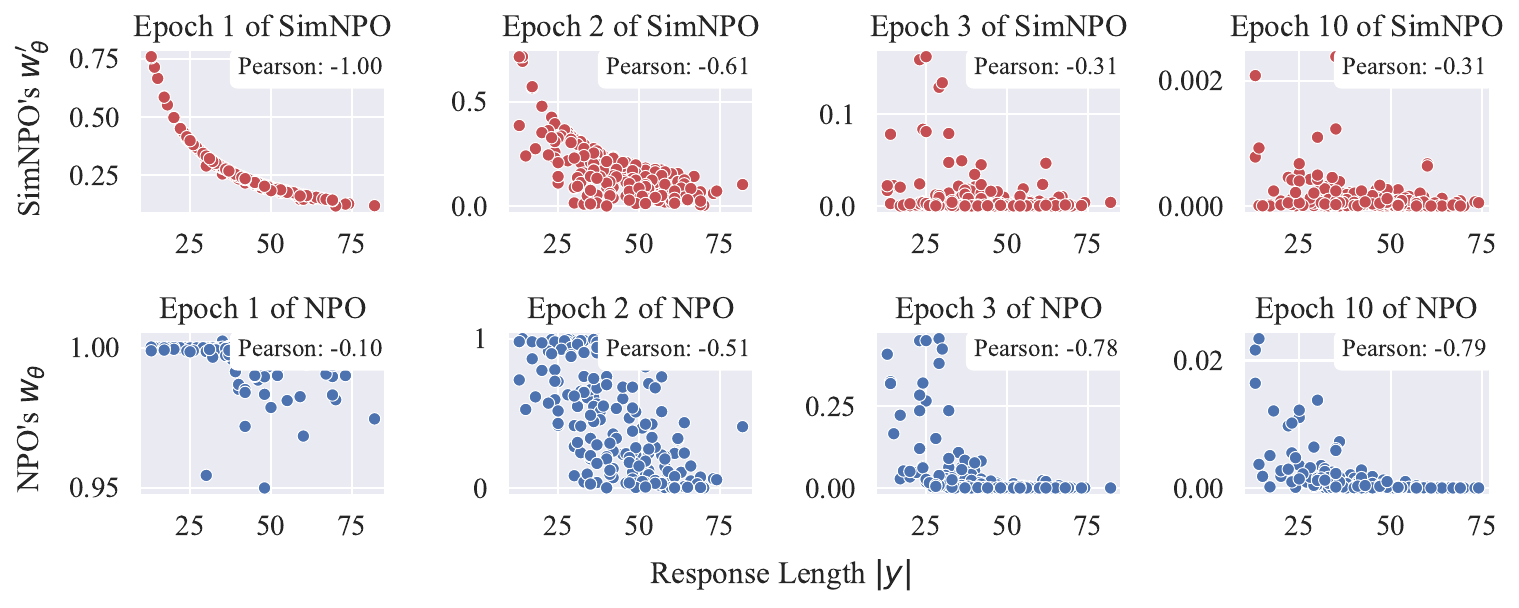}
\vspace*{-4mm}
\caption{
\small{Gradient weight smoothing of NPO ($w_\btheta$) and SimNPO ($w_\btheta^\prime$) vs. forget data response length $|y|$ across different epochs (1, 2, 3, and 10) on TOFU Forget05. The Pearson correlation in the upper right corner indicates the relationship between gradient weight smoothing and response length. The SimNPO's weights $w_\btheta^\prime$ have been rescaled (by $\times 10$) for ease of visualization.}
}
\vspace{-1mm}
\label{fig: illustration_weight_smoothing_SimNPO}
\end{figure}



(b) In addition, the reference-free, length-normalized weight smoothing prevents early-stage ineffectiveness during unlearning.  It can be shown from \eqref{eq: grad_SimNPO_Nogamma}  that $w^\prime_{\btheta} (x,y) < {2}/{|y|}$, with the distribution of weights $w^\prime_{\btheta} (x,y)$ depending on the specific forget data samples. This contrasts with NPO, where the weight distribution concentrated around $w_{\btheta} (x,y) \approx 1$ during the early unlearning stage.
Extended from Fig.\,\ref{fig: NPO_grad_weights}-(a)\&(b),  \textbf{Fig.\,\ref{fig: illustration_weight_smoothing_SimNPO}} provides a detailed comparison between the gradient weights of SimNPO and NPO. We find that SimNPO tends to prioritize short-length forget data that are initially harder to forget during the first two unlearning epochs. At later epochs, the gradient weights become more uniform, reflecting that SimNPO can then treat different forget data with even optimization power. This trend is different from  NPO, which assigns more uniform gradient weights early on and starts to account for data-specific difficulty only in the later stages of unlearning. 
Besides the above advantage, we also find that SimNPO's new weight smoothing scheme does not compromise the overall unlearning speed compared to NPO. This is supported by the divergence rate from the pre-trained state shown in \textbf{Fig.\,\ref{fig: plot_divergence}} and our theoretical discussion in \textbf{Appendix\,\ref{app: explanation_divergence}}.

\begin{wrapfigure}{r}{.49\textwidth}
\vspace*{-4mm}
    \centering
    \begin{tabular}{cc} 
       \hspace*{-2mm} \includegraphics[width=0.23\textwidth]{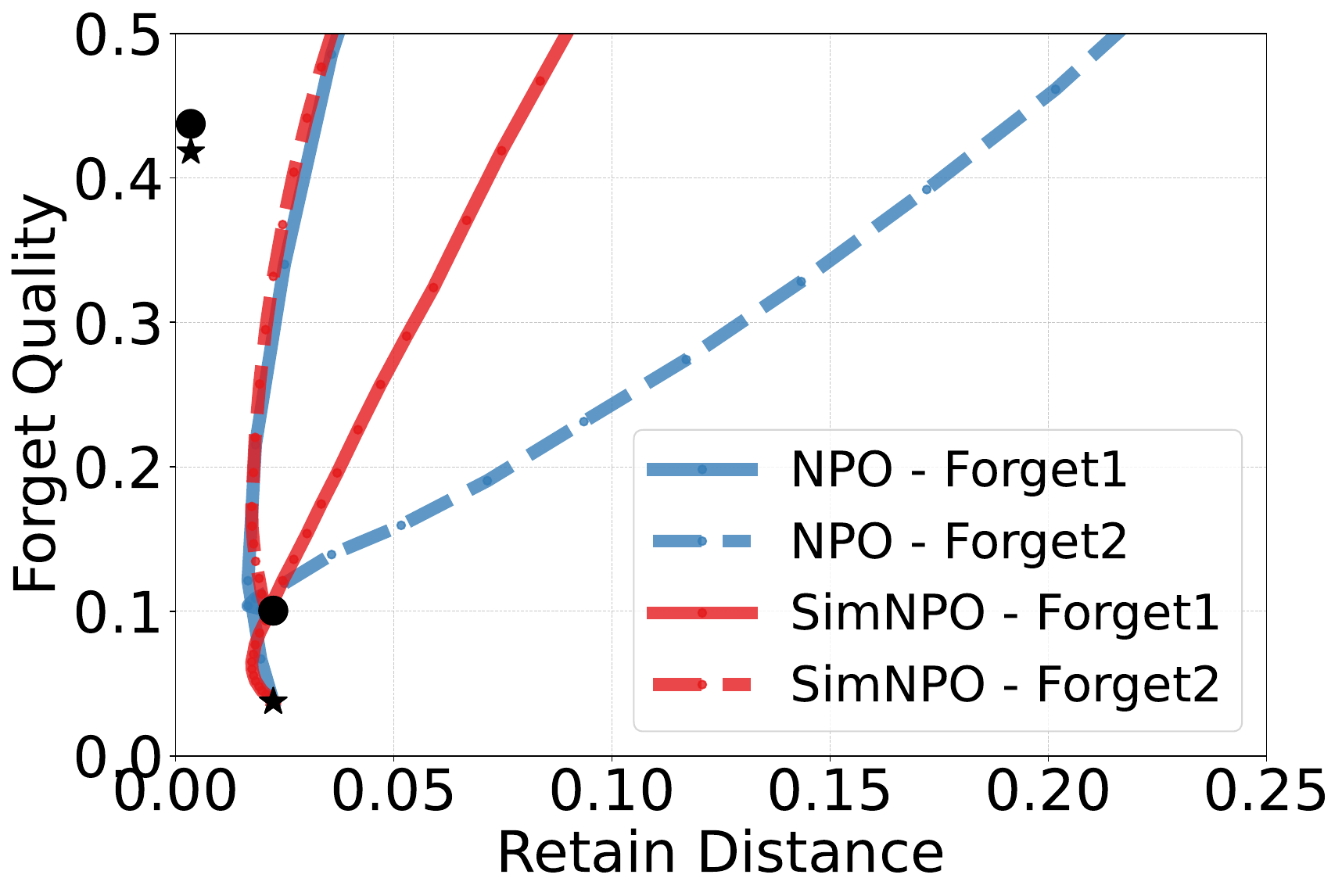} &
       \hspace*{-6mm} 
        \includegraphics[width=0.23\textwidth]{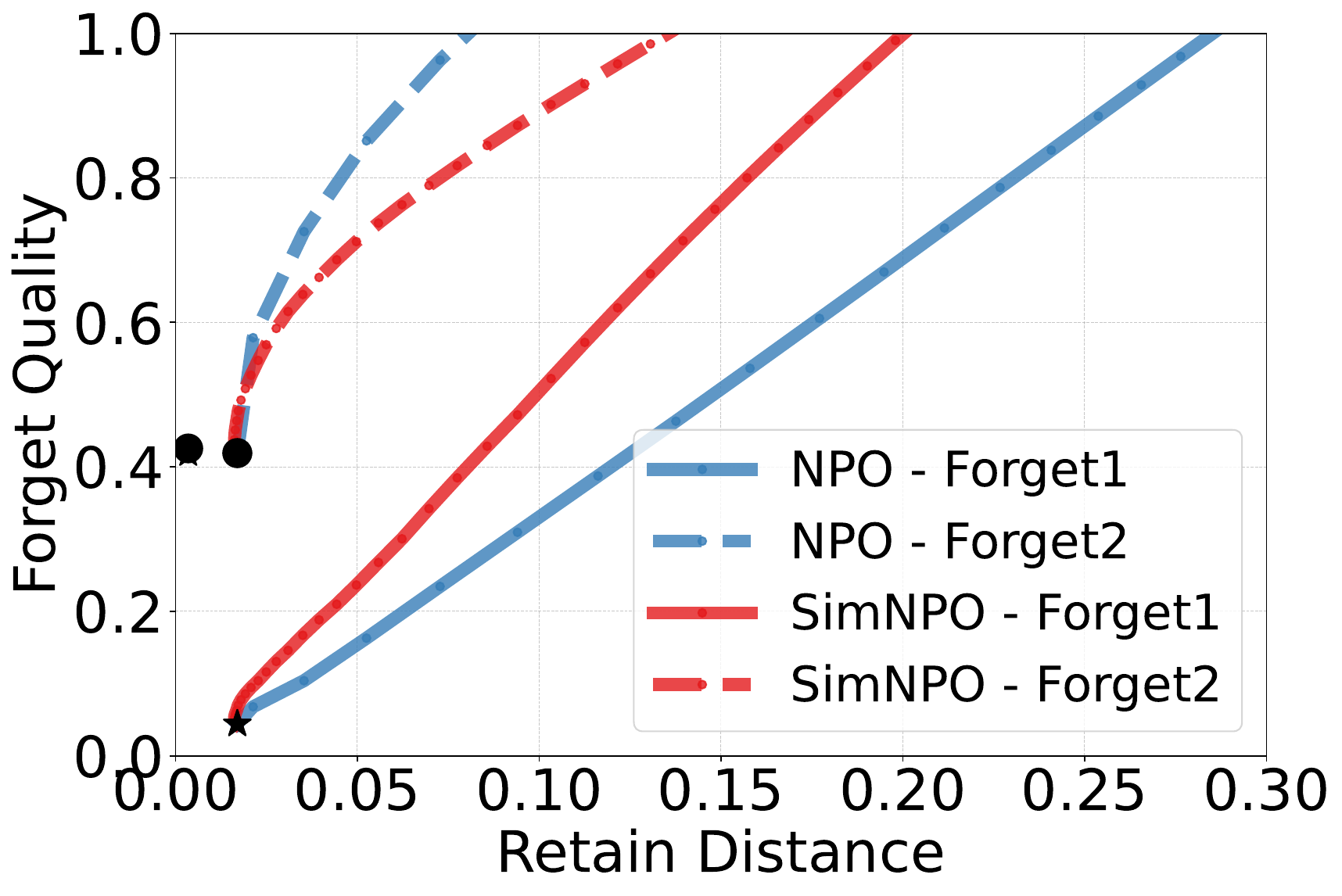} \\
      \hspace*{-2mm}  \small{(a) Different length.} &
        \hspace*{-6mm}  \small{(b) Different memorization.}
    \end{tabular}
    \vspace*{-3mm}
    \caption{\small{Tradeoffs between forget quality (higher $\uparrow$ is better) and retain distance (lower $\downarrow$ is better) along the unlearning path of NPO and SimNPO in the synthetic experiments. The symbols \((\star, \bullet)\) near the $y$-axis of both figures indicate the performance of the retrained model on Forget1 and Forget2, respectively.}
    }
    \label{fig:synthetic_mc_1}
\vspace*{-3.5mm}
\end{wrapfigure}

\textbf{Further analyses via a mixture of Markov chains.} In addition to the above insights, we further validate SimNPO's advantages to overcome NPO's limitations (Sec.\,\ref{sec: NPO_limitations}) using a synthetic setup. For ease of controlling the unlearning difficulties of different forget data points,  
we consider the problem of unlearning
on a mixture of Markov chains with a state space of size 10 ($s=1,\ldots,10$). The \textit{retain distribution} consists of Markov chains that transition uniformly among states $\{1,2,3\}$. The \textit{forget distribution} is a mixture of two components: \emph{Forget1}, where the chains transition uniformly among $\{4,5,6\}$, and \emph{Forget2}, where they move uniformly among $\{7,8,9\}$. A small leakage probability allows the chains to transition outside their designated states occasionally, including state $10$, which is not a designated state for any of the chains. We generate 10,000 samples for the retain distribution and 5,000 samples each for Forget1 and Forget2. A GPT-2 model is pretrained on these samples and serves as the initial model. We apply NPO and SimNPO to unlearn the forget distributions. Forget and retain performance is evaluated using the KL-divergence between  predicted and true transition probabilities of the Markov chains. See \textbf{Appendix\,\ref{appx: synthetic}} for details.  We present our results in \textbf{Fig.\,\ref{fig:synthetic_mc_1}} and summarize the insights below.

\emph{SimNPO achieves more balanced unlearning across data of varying lengths compared to NPO.}
To validate this, we set the retain distribution and Forget1 with a sequence length of 20, while Forget2 is assigned a shorter sequence length of 5, representing a mix of long and short responses.
 \textbf{Fig.\,\ref{fig:synthetic_mc_1}\,(a)} shows that NPO exhibits a worse tradeoff between retain distance and forget quality on short responses  (\textit{i.e.}, Forget2) compared with SimNPO. That is, to achieve the same forget quality on Forget2 as the retrained model (with forget quality $0.44$), NPO incurs a higher retain distance than SimNPO.  As a result, NPO has an overall larger retain distance when unlearning the entire Forget distribution. In contrast, SimNPO shows more consistent performance across Forget1 and Forget2, with less variance in its tradeoff.

\emph{SimNPO achieves more balanced unlearning across data of varying memorization compared to NPO.}
In the second case,  we set the retain distribution, Forget1 and Forget2 all with a sequence length of $20$. However, we exclude Forget2 during pretraining. This setup simulates a scenario where the initial model (\textit{i.e.}, the reference model in NPO) exhibits varying levels of memorization for the forget data: strongly memorized  Forget1 against Forget2. \textbf{Fig.\,\ref{fig:synthetic_mc_1}\,(b)} shows that NPO exhibits a larger gap between Forget1 and Forget2 for the same Retain distance, leading to over-unlearning weakly-memorized data (as shown by the comparison between NPO-Forget2 vs. SimNPO-Forget2) and under-unlearning strongly-memorized data (as shown by the comparison between NPO-Forget1 vs. SimNPO-Forget1). SimNPO has a better balance during unlearning across data with varying levels of memorization.

\vspace*{-1mm}
\section{Experiments}
\label{sec: exps}

\vspace*{-1mm}
\subsection{Experiment setups}
\vspace*{-2mm}

\noindent \textbf{Datasets and methods.} We evaluate unlearning tasks on three benchmark datasets: TOFU \citep{maini2024tofu}, MUSE \citep{shi2024muse}, and WMDP \citep{li2024wmdp}. TOFU includes `{Forget05}' and `{Forget10}' scenarios, representing 5\% and 10\% forget sets, respectively. MUSE focuses on `{Books}' and `{News}' forgetting scenarios, while WMDP targets knowledge-based unlearning of hazardous biosecurity information.

\noindent \textbf{LLM unlearning methods and evaluation.} We evaluate a range of unlearning methods, including \textbf{Retrain}, \textbf{SimNPO}, \textbf{NPO}, \textbf{GA}, and \textbf{GradDiff}. In addition, we incorporate several task-specific approaches: the rejection-based method \textbf{IDK}, which replaces positive responses in DPO with generic answers such as “I don’t know”~\cite{maini2024tofu}, and \textbf{RKLD}~\cite{wang2024rkld} in the TOFU; the \textbf{Task Vector} method used in MUSE~\cite{shi2024muse}; and the representation misdirection unlearning method \textbf{RMU} in WMDP~\cite{li2024wmdp}. Evaluation metrics for each benchmark are summarized in Table\,\ref{tab:tasks_evaluations} and further detailed in Appendix\,\ref{app: exp_setup}. For the relearning attack, we use 20\% of the TOFU Forget05 set and retrain over three epochs. Please refer to \textbf{Appendix\,\ref{app: exp_setup}} for full experimental details.

\vspace*{-3mm}
\subsection{Experiment results}
\vspace*{-2mm}

\begin{table*}[htb]
\begin{center}
\caption{\small{Unlearning performance on TOFU Forget05 using the LLaMA2-7B-chat model. `Prob.' indicates the probability metrics, as summarized in Table\,\ref{tab:tasks_evaluations}, with forget quality (FQ) and model utility (MU) serving as the primary metrics.  Results are averaged over five random trials. The best FQ and MU are highlighted in \textbf{bold}.}
}
\vspace*{2mm}
\resizebox{0.99\textwidth}{!}{
\begin{tabular}{c|ccc|c|ccccccccc|c}
\toprule[1pt]
\midrule
\multicolumn{1}{c|}{\multirow{3}{*}{\textbf{Method}}} & \multicolumn{4}{c|}{\textbf{Unlearning Efficacy}} & \multicolumn{10}{c}{\textbf{Utility Preservation}} \\
\cmidrule{2-15}
& \multicolumn{3}{c|}{Forget Set}  && \multicolumn{3}{c}{Real Authors} & \multicolumn{3}{c}{World Facts} & \multicolumn{3}{c|}{Retain Set} &  \\
&{(1-Rouge-L)\textcolor{black}{\textuparrow}}                        & {(1-Prob.)\textcolor{black}{\textuparrow}} & Truth ratio\textcolor{black}{\textuparrow}&{\multirow{-2}{*}{{FQ\textuparrow}}}    & Rouge-L\textcolor{black}{\textuparrow} &Prob.\textcolor{black}{\textuparrow}& Truth ratio\textcolor{black}{\textuparrow}  & Rouge-L\textcolor{black}{\textuparrow} &Prob.\textcolor{black}{\textuparrow}& Truth ratio\textcolor{black}{\textuparrow} & Rouge-L\textcolor{black}{\textuparrow}&Prob.\textcolor{black}{\textuparrow}& Truth ratio\textcolor{black}{\textuparrow} & \multirow{-2}{*}{MU\textcolor{black}{\textuparrow}}\\ 
\midrule
{Original} & 0.04 & 0.01 & 0.49 & 0.00     & 0.93 & 0.44 & 0.58 & 0.91 & 0.43 & 0.55 & 0.98 & 0.99 & 0.48 & 0.62 \\
{Retrain}  & 0.61 & 0.85 & 0.66 & 1.00     & 0.92 & 0.44 & 0.57 & 0.90 & 0.43 & 0.54 & 0.97 & 0.99 & 0.48 & 0.62 \\
\midrule
{GA}       & 1.00 & 1.00 & 0.66 & 1.9e-9 & 0.00 & 0.20 & 0.40 & 0.00 & 0.30 & 0.28 & 0.00 & 0.00 & 0.15 & 0.00 \\
{GradDiff} & 1.00 & 1.00 & 0.60 & 3.6e-9 & 0.59 & 0.59 & 0.81 & 0.88 & 0.46 & 0.59 & 0.42 & 0.49 & 0.48 & 0.56 \\
{IDK}      & 0.98 & 0.40 & 0.55 & 1.9e-9 & 0.65 & 0.48 & 0.63 & 0.82 & 0.44 & 0.55 & 0.55 & 0.86 & 0.43 & 0.57 \\
{RKLD}     & 0.69 & 0.96 & 0.66 & 0.79     & 0.92 & 0.47 & 0.61 & 0.87 & 0.47 & 0.58 & 0.58 & 0.52 & 0.43 & 0.56 \\
{NPO}      & 0.73 & 0.94 & 0.67 & 0.79     & 0.91 & 0.50 & 0.62 & 0.90 & 0.50 & 0.61 & 0.47 & 0.51 & 0.44 & 0.57 \\
\midrule
\rowcolor{Gray}
\textbf{SimNPO}   & 0.74 & 0.97 & 0.69 & \textbf{0.99}     & 0.90 & 0.50 & 0.64 & 0.90 & 0.48 & 0.60 & 0.54 & 0.56 & 0.44 & \textbf{0.58} \\
\midrule
\bottomrule
\end{tabular}
}
\label{tab: tofu}
\vspace*{-6mm}
\end{center}
\end{table*}

\noindent \textbf{Performance on TOFU.}
In \textbf{Table\,\ref{tab: tofu}}, we present the unlearning performance of SimNPO and its various baselines on TOFU Forget05, covering both effectiveness metrics and utility  metrics as shown in Table\,\ref{tab:tasks_evaluations}. 
`{FQ}' stands for forget quality, and `{MU}' represents model utility.
These two metrics serve as the primary performance indicators for LLM unlearning on  TOFU. 
SimNPO outperforms NPO in both FQ and MU, and is the closest approximate unlearning method to Retrain. Except for NPO and RKLD, the other unlearning baselines (GA, GradDiff, and IDK) are not effective, as implied by their FQ values being smaller than $0.01$, where FQ indicates the $p$-value for rejecting the indistinguishability between the unlearned model and Retrain on  TOFU.
 In \textbf{Table\,\ref{tab: examples_tofu_extended} of Appendix\,\ref{app: more_exam}}, we also provide examples of model responses after unlearning using SimNPO, Retrain, and NPO, along with label to degenerate. We observe that, in some cases (\textit{e.g.}, responses against the input queries Q1 and Q2 in Table\,\ref{tab: examples_tofu_extended}), the NPO-unlearned model generates \textit{repeated texts} in response. While this repetition does not reveal the information intended for unlearning, it differs noticeably from Retrain. In contrast, SimNPO produces unlearning responses more closely aligned with those generated by Retrain. More results on TOFU Forget10 are in \textbf{Table\,\ref{tab: tofu_forget10} of \textbf{Appendix\,\ref{app: exp_tofu}}}.

\begin{wraptable}{r}{0.435\textwidth}
\begin{center}
\vspace{-8mm}
\caption{\small{
Performance of various unlearning methods on MUSE News (LLaMA2-7B) and MUSE Books (ICLM-7B).
} 
}
\resizebox{0.435\textwidth}{!}{
\begin{tabular}{c|ccc|c}
\toprule[1pt]
\midrule
\multicolumn{1}{c|}{\multirow{3}{*}{\textbf{Method}}} & \multicolumn{3}{c|}{\textbf{Unlearning Efficacy}} & \multicolumn{1}{c}{\textbf{Utility}} \\
\cmidrule{2-5}
& \begin{tabular}{c}
     VerbMem   \\
     $\mathcal{D}_f$ ($\downarrow$)
\end{tabular}

 & \begin{tabular}{c}
   KnowMem \\
      $\mathcal{D}_f$ ($\downarrow$)
\end{tabular}  & \begin{tabular}{c}
    PrivLeak \\
    ($\rightarrow$ 0)
\end{tabular}  & \begin{tabular}{c}
     KnowMem  \\
    $\mathcal{D}_r$ ($\uparrow$)
\end{tabular} \\ 
\midrule
\multicolumn{5}{c}{\textbf{MUSE News}} \\\midrule
{Original} & 58.29  & 62.93  & -98.71    & 54.31  \\
{Retrain }          & 20.75 & 33.32 & 0.00    & 53.79 \\
\midrule
{GA}               & 0.00  & 0.00  & 20.14   & 0.00  \\
{GradDiff}          & 4.85  & 31.29  & 108.12   & 28.21  \\
{Task Vector}       & 77.42 & 58.76 & -100.00 & 47.94 \\
{NPO}               & 2.53  & 56.93 & 108.91  & 37.58 \\
\midrule
\rowcolor{Gray}
\textbf{SimNPO}            & 2.34  & 44.84  & 72.93  & 39.65 \\
\midrule
\multicolumn{5}{c}{\textbf{MUSE Books}} \\\midrule
{Original} & 99.56  & 58.32  & -56.32   & 67.01  \\
{Retrain}           & 14.30 & 28.90 & 0.00   & 74.50 \\
\midrule
{GA}                & 0.00  & 0.00  & -24.07 & 0.00  \\
{GradDiff}          & 0.00  & 0.00  & -24.59 & 0.13  \\
{Task Vector}       & 99.31 & 35.55 & -83.78 & 62.55 \\
{NPO}               & 0.00  & 0.00  & -31.17 & 23.71 \\
\midrule
\rowcolor{Gray}
\textbf{SimNPO}            & 0.00  & 0.00  & -19.82  & 48.27 \\
\midrule
\bottomrule
\end{tabular}
}
\label{tab: muse}
\vspace*{-5mm}
\end{center}
\end{wraptable}
\noindent\textbf{Performance on MUSE and WMDP.}
\textbf{Table\,\ref{tab: muse}} compares SimNPO with other methods, on MUSE News and Books, with evaluation metrics in Table\,\ref{tab:tasks_evaluations}. Compared to NPO, SimNPO preserves higher utility while achieving stronger unlearning. On $\mathcal{D}_r$, KnowMem is 39.65 (News) and 48.27 (Books), while on $\mathcal{D}_f$, it is 44.84 (News) and 0.00 (Books). SimNPO also attains a PrivLeak value closer to 0 than NPO (72.93 for News, $-31.17$ for Books), indicating it better approximates complete data removal \citep{shi2024muse}. Compared to other methods, SimNPO strikes the best balance between utility and unlearning. We further evaluate sequential unlearning on MUSE News (\textbf{Fig.\,\ref{fig: muse_seq}} in \textbf{Appendix\,\ref{app: exp_muse}}), where SimNPO consistently outperforms NPO as requests increase. 
Due to space constraints, we present SimNPO's performance on the WMDP dataset in \textbf{Appendix\,\ref{app: exp_wmdp}}.

\begin{wrapfigure}{r}{0.25\textwidth}
    \centering
    \vspace*{-4mm}
    \hspace*{6mm} 
\includegraphics[width=0.2\textwidth]{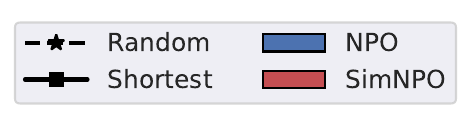}\\
    \includegraphics[width=\linewidth]{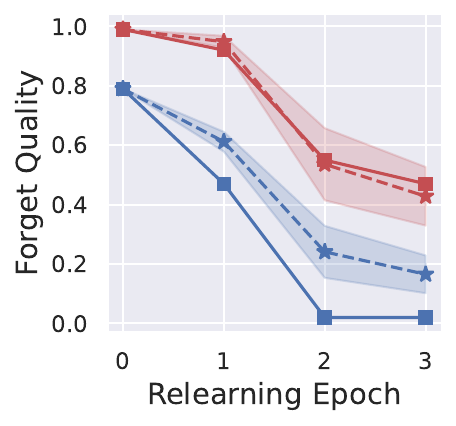}
    \vspace*{-8mm}
    \caption{\small{NPO and SimNPO under random/shortest relearn attack vs. epochs on TOFU Forget05.}}
    \label{fig: relearn}
    \vspace*{-3mm}
\end{wrapfigure}

\noindent \textbf{Unlearning robustness against length-variant relearning attacks.} Recent studies \citep{lynch2024eight,hu2024jogging} show that unlearning methods are vulnerable to relearning attacks, where forgotten information can be recovered by finetuning on a subset of the forget set. We evaluate SimNPO’s robustness against such attacks, showing it to outperform NPO, especially for short-length response data.
\textbf{Fig.\,\ref{fig: relearn}} presents the forget quality of SimNPO and NPO under relearning attacks against the number of relearning epochs. Relearning is performed on the forget subset, which is either the shortest 20\% of responses from the TOFU Forget05 dataset or an equal-size random subset. We refer to these attacks as `shortest-relearn' and `random-relearn', respectively. The random-relearn case is conducted 5 times, with both average robustness and variance in {Fig.\,\ref{fig: relearn}}.
As we can see, SimNPO demonstrates improved robustness over NPO, evidenced by higher forget quality and a slower decline in forget quality as the relearning epoch increases. NPO is less robust against the shortest-relearn attack compared to the random-relearn attack. In contrast, SimNPO is resilient to both types of relearning. This is expected since SimNPO addresses the limitation (L1), as explained in Sec.\,\ref{sec: NPO_limitations}.




\vspace*{-3mm}
\section{Conclusion}
\vspace*{-2mm}

We identified a reference model bias in negative preference optimization (NPO) that limits unlearning effectiveness. To address this, we proposed SimNPO, a simpler framework leveraging preference optimization without a reference model. SimNPO consistently outperforms NPO across standard benchmarks such as TOFU, MUSE, and WMDP, and demonstrates additional advantages in unlearning robustness and the application to reasoning model unlearning. 



\section*{Broader Impact}
\label{appx: impact}
On the positive side, we have demonstrated the utility of preference optimization in machine unlearning. This connection enables more efficient unlearning operations in LLMs, improving data privacy protections and supporting compliance with regulatory requirements. Additionally, given the relationship between preference optimization and model editing, our work encourages further exploration in these areas, contributing to the development of models that are 
easier to customize and become safer to deploy.
On the negative side, the methods we developed could be misused to selectively erase ``essential'' (rather than ``unwanted'') concepts or knowledge, raising ethical and legal concerns. To mitigate this risk, it is essential to ensure that unlearning applications adhere to strict ethical guidelines to prevent misuse. We hope our research fosters the development of safe, reliable, and human-aligned LLMs.

\section*{Limitations}
\label{appx: limitations}

While SimNPO mitigates the reference model bias present in NPO and improves gradient weight smoothing to better adjust divergence speed based on the varying unlearning difficulties of forget data samples, both frameworks still rely on promoting divergence to achieve unlearning. This reliance inevitably results in some degree of utility loss.
This limitation becomes especially evident in knowledge unlearning or model capability removal scenarios, such as in the WMDP unlearning benchmark. Consequently, SimNPO has yet to fully resolve the challenge of balancing unlearning effectiveness with  model utility. {Additionally, establishing theoretical guarantees for SimNPO remains an important area for future research.}



\section*{Acknowledgement}
C. Fan, J. Liu, J. Jia, and S. Liu were supported 
in part by the National Science Foundation (NSF) CISE Core Program Awards IIS-2207052 and IIS-2504263, the NSF CAREER Award IIS-2338068, the ARO Award W911NF2310343, the Amazon Research Award for AI in Information Security, the Open Philanthropy Research Award, and the Center for AI Safety (CAIS) Compute Award. We also extend our gratitude to the MIT-IBM Watson AI Lab, IBM Research for their support in this project. 
 L. Lin, R. Zhang, and S. Mei were supported in part by NSF CCF-2315725, NSF Career DMS-2339904, ONR N00014-24-S-B001, an Amazon Research Award, and a Google Research Scholar Award. We also thank the support from the Center for AI Safety Compute Cluster. Finally, we express our appreciation to Yuguang Yao for his help in figure plotting.



\bibliography{refs/MU,refs/MU_SLiu,refs/PO}

\begin{thebibliography}{67}
\providecommand{\natexlab}[1]{#1}
\providecommand{\url}[1]{#1}
\csname url@samestyle\endcsname
\providecommand{\newblock}{\relax}
\providecommand{\bibinfo}[2]{#2}
\providecommand{\BIBentrySTDinterwordspacing}{\spaceskip=0pt\relax}
\providecommand{\BIBentryALTinterwordstretchfactor}{4}
\providecommand{\BIBentryALTinterwordspacing}{\spaceskip=\fontdimen2\font plus
\BIBentryALTinterwordstretchfactor\fontdimen3\font minus \fontdimen4\font\relax}
\providecommand{\BIBforeignlanguage}[2]{{%
\expandafter\ifx\csname l@#1\endcsname\relax
\typeout{** WARNING: IEEEtranN.bst: No hyphenation pattern has been}%
\typeout{** loaded for the language `#1'. Using the pattern for}%
\typeout{** the default language instead.}%
\else
\language=\csname l@#1\endcsname
\fi
#2}}
\providecommand{\BIBdecl}{\relax}
\BIBdecl

\bibitem[Huang et~al.(2024)Huang, Sun, Wang, Wu, Zhang, Li, Gao, Huang, et~al.]{sun2024trustllm}
Y.~Huang, L.~Sun, H.~Wang, S.~Wu, Q.~Zhang, Y.~Li, C.~Gao, Y.~Huang \emph{et~al.}, ``Position: {T}rust{LLM}: Trustworthiness in large language models,'' in \emph{Proceedings of the 41st International Conference on Machine Learning}, ser. Proceedings of Machine Learning Research, vol. 235, 21--27 Jul 2024, pp. 20\,166--20\,270.

\bibitem[Wang et~al.(2023)Wang, Chen, Pei, Xie, Kang, Zhang, Xu, Xiong, Dutta, Schaeffer, et~al.]{wang2023decodingtrust}
B.~Wang, W.~Chen, H.~Pei, C.~Xie, M.~Kang, C.~Zhang, C.~Xu, Z.~Xiong, R.~Dutta, R.~Schaeffer \emph{et~al.}, ``Decodingtrust: A comprehensive assessment of trustworthiness in gpt models.'' in \emph{NeurIPS}, 2023.

\bibitem[Li et~al.(2024)Li, Pan, Gopal, Yue, Berrios, Gatti, Li, Dombrowski, Goel, Phan, et~al.]{li2024wmdp}
N.~Li, A.~Pan, A.~Gopal, S.~Yue, D.~Berrios, A.~Gatti, J.~D. Li, A.-K. Dombrowski, S.~Goel, L.~Phan \emph{et~al.}, ``The wmdp benchmark: Measuring and reducing malicious use with unlearning,'' \emph{arXiv preprint arXiv:2403.03218}, 2024.

\bibitem[Shi et~al.(2024)Shi, Lee, Huang, Malladi, Zhao, Holtzman, Liu, Zettlemoyer, Smith, and Zhang]{shi2024muse}
W.~Shi, J.~Lee, Y.~Huang, S.~Malladi, J.~Zhao, A.~Holtzman, D.~Liu, L.~Zettlemoyer, N.~A. Smith, and C.~Zhang, ``Muse: Machine unlearning six-way evaluation for language models,'' \emph{arXiv preprint arXiv:2407.06460}, 2024.

\bibitem[Liu et~al.(2025)Liu, Yao, Jia, Casper, Baracaldo, Hase, Yao, Liu, Xu, Li, et~al.]{liu2024rethinking}
S.~Liu, Y.~Yao, J.~Jia, S.~Casper, N.~Baracaldo, P.~Hase, Y.~Yao, C.~Y. Liu, X.~Xu, H.~Li \emph{et~al.}, ``Rethinking machine unlearning for large language models,'' \emph{Nature Machine Intelligence}, pp. 1--14, 2025.

\bibitem[Rosen(2011)]{rosen2011right}
J.~Rosen, ``The right to be forgotten,'' \emph{Stan. L. Rev. Online}, vol.~64, p.~88, 2011.

\bibitem[Hoofnagle et~al.(2019)Hoofnagle, van~der Sloot, and Borgesius]{hoofnagle2019european}
C.~J. Hoofnagle, B.~van~der Sloot, and F.~Z. Borgesius, ``The european union general data protection regulation: what it is and what it means,'' \emph{Information \& Communications Technology Law}, vol.~28, no.~1, pp. 65--98, 2019.

\bibitem[Cao and Yang(2015)]{cao2015towards}
Y.~Cao and J.~Yang, ``Towards making systems forget with machine unlearning,'' in \emph{2015 IEEE symposium on security and privacy}.\hskip 1em plus 0.5em minus 0.4em\relax IEEE, 2015, pp. 463--480.

\bibitem[Warnecke et~al.(2021)Warnecke, Pirch, Wressnegger, and Rieck]{warnecke2021machine}
A.~Warnecke, L.~Pirch, C.~Wressnegger, and K.~Rieck, ``Machine unlearning of features and labels,'' \emph{arXiv preprint arXiv:2108.11577}, 2021.

\bibitem[Bourtoule et~al.(2021)Bourtoule, Chandrasekaran, Choquette-Choo, Jia, Travers, Zhang, Lie, and Papernot]{bourtoule2021machine}
L.~Bourtoule, V.~Chandrasekaran, C.~A. Choquette-Choo, H.~Jia, A.~Travers, B.~Zhang, D.~Lie, and N.~Papernot, ``Machine unlearning,'' in \emph{2021 IEEE Symposium on Security and Privacy (SP)}.\hskip 1em plus 0.5em minus 0.4em\relax IEEE, 2021, pp. 141--159.

\bibitem[Thudi et~al.(2022)Thudi, Deza, Chandrasekaran, and Papernot]{thudi2022unrolling}
A.~Thudi, G.~Deza, V.~Chandrasekaran, and N.~Papernot, ``Unrolling sgd: Understanding factors influencing machine unlearning,'' in \emph{2022 IEEE 7th European Symposium on Security and Privacy (EuroS\&P)}.\hskip 1em plus 0.5em minus 0.4em\relax IEEE, 2022, pp. 303--319.

\bibitem[Kurmanji et~al.(2024)Kurmanji, Triantafillou, Hayes, and Triantafillou]{kurmanji2024towards}
M.~Kurmanji, P.~Triantafillou, J.~Hayes, and E.~Triantafillou, ``Towards unbounded machine unlearning,'' \emph{Advances in neural information processing systems}, vol.~36, 2024.

\bibitem[Jia et~al.(2023)Jia, Liu, Ram, Yao, Liu, Liu, Sharma, and Liu]{jia2023model}
J.~Jia, J.~Liu, P.~Ram, Y.~Yao, G.~Liu, Y.~Liu, P.~Sharma, and S.~Liu, ``Model sparsity can simplify machine unlearning,'' in \emph{Thirty-seventh Conference on Neural Information Processing Systems}, 2023.

\bibitem[Gandikota et~al.(2023)Gandikota, Materzynska, Fiotto-Kaufman, and Bau]{gandikota2023erasing}
R.~Gandikota, J.~Materzynska, J.~Fiotto-Kaufman, and D.~Bau, ``Erasing concepts from diffusion models,'' in \emph{Proceedings of the IEEE/CVF International Conference on Computer Vision}, 2023, pp. 2426--2436.

\bibitem[Fan et~al.(2024{\natexlab{a}})Fan, Liu, Zhang, Wei, Wong, and Liu]{fan2023salun}
C.~Fan, J.~Liu, Y.~Zhang, D.~Wei, E.~Wong, and S.~Liu, ``Salun: Empowering machine unlearning via gradient-based weight saliency in both image classification and generation,'' in \emph{International Conference on Learning Representations}, 2024.

\bibitem[Eldan and Russinovich(2023)]{eldan2023whos}
R.~Eldan and M.~Russinovich, ``Who's harry potter? approximate unlearning in llms,'' 2023.

\bibitem[Yao et~al.(2023)Yao, Xu, and Liu]{yao2023large}
Y.~Yao, X.~Xu, and Y.~Liu, ``Large language model unlearning,'' \emph{arXiv preprint arXiv:2310.10683}, 2023.

\bibitem[Maini et~al.(2024)Maini, Feng, Schwarzschild, Lipton, and Kolter]{maini2024tofu}
P.~Maini, Z.~Feng, A.~Schwarzschild, Z.~C. Lipton, and J.~Z. Kolter, ``Tofu: A task of fictitious unlearning for llms,'' 2024.

\bibitem[Zhang et~al.(2024{\natexlab{a}})Zhang, Lin, Bai, and Mei]{zhang2024negative}
R.~Zhang, L.~Lin, Y.~Bai, and S.~Mei, ``Negative preference optimization: From catastrophic collapse to effective unlearning,'' \emph{arXiv preprint arXiv:2404.05868}, 2024.

\bibitem[Jia et~al.(2024)Jia, Zhang, Zhang, Liu, Runwal, Diffenderfer, Kailkhura, and Liu]{jia2024soul}
J.~Jia, Y.~Zhang, Y.~Zhang, J.~Liu, B.~Runwal, J.~Diffenderfer, B.~Kailkhura, and S.~Liu, ``Soul: Unlocking the power of second-order optimization for llm unlearning,'' \emph{arXiv preprint arXiv:2404.18239}, 2024.

\bibitem[Liu et~al.(2022{\natexlab{a}})Liu, Liu, and Stone]{liu2022continual}
B.~Liu, Q.~Liu, and P.~Stone, ``Continual learning and private unlearning,'' in \emph{Conference on Lifelong Learning Agents}.\hskip 1em plus 0.5em minus 0.4em\relax PMLR, 2022, pp. 243--254.

\bibitem[Rafailov et~al.(2024)Rafailov, Sharma, Mitchell, Manning, Ermon, and Finn]{rafailov2024direct}
R.~Rafailov, A.~Sharma, E.~Mitchell, C.~D. Manning, S.~Ermon, and C.~Finn, ``Direct preference optimization: Your language model is secretly a reward model,'' \emph{Advances in Neural Information Processing Systems}, vol.~36, 2024.

\bibitem[Meng et~al.(2024)Meng, Xia, and Chen]{meng2024simpo}
Y.~Meng, M.~Xia, and D.~Chen, ``Sim{PO}: Simple preference optimization with a reference-free reward,'' in \emph{The Thirty-eighth Annual Conference on Neural Information Processing Systems}, 2024.

\bibitem[Lynch et~al.(2024)Lynch, Guo, Ewart, Casper, and Hadfield-Menell]{lynch2024eight}
A.~Lynch, P.~Guo, A.~Ewart, S.~Casper, and D.~Hadfield-Menell, ``Eight methods to evaluate robust unlearning in llms,'' \emph{arXiv preprint arXiv:2402.16835}, 2024.

\bibitem[Hu et~al.(2024)Hu, Fu, Wu, and Smith]{hu2024jogging}
S.~Hu, Y.~Fu, Z.~S. Wu, and V.~Smith, ``Jogging the memory of unlearned model through targeted relearning attack,'' \emph{arXiv preprint arXiv:2406.13356}, 2024.

\bibitem[Nguyen et~al.(2022)Nguyen, Huynh, Nguyen, Liew, Yin, and Nguyen]{nguyen2022survey}
T.~T. Nguyen, T.~T. Huynh, P.~L. Nguyen, A.~W.-C. Liew, H.~Yin, and Q.~V.~H. Nguyen, ``A survey of machine unlearning,'' \emph{arXiv preprint arXiv:2209.02299}, 2022.

\bibitem[Triantafillou et~al.(2024)Triantafillou, Kairouz, Pedregosa, Hayes, Kurmanji, Zhao, Dumoulin, Junior, Mitliagkas, Wan, et~al.]{triantafillou2024we}
E.~Triantafillou, P.~Kairouz, F.~Pedregosa, J.~Hayes, M.~Kurmanji, K.~Zhao, V.~Dumoulin, J.~J. Junior, I.~Mitliagkas, J.~Wan \emph{et~al.}, ``Are we making progress in unlearning? findings from the first neurips unlearning competition,'' \emph{arXiv preprint arXiv:2406.09073}, 2024.

\bibitem[Liu et~al.(2022{\natexlab{b}})Liu, Fan, Chen, Liu, Ma, Wang, and Ma]{liu2022backdoor}
Y.~Liu, M.~Fan, C.~Chen, X.~Liu, Z.~Ma, L.~Wang, and J.~Ma, ``Backdoor defense with machine unlearning,'' in \emph{IEEE INFOCOM 2022-IEEE Conference on Computer Communications}.\hskip 1em plus 0.5em minus 0.4em\relax IEEE, 2022, pp. 280--289.

\bibitem[Fan et~al.(2024{\natexlab{b}})Fan, Liu, Hero, and Liu]{fan2024challenging}
C.~Fan, J.~Liu, A.~Hero, and S.~Liu, ``Challenging forgets: Unveiling the worst-case forget sets in machine unlearning,'' \emph{arXiv preprint arXiv:2403.07362}, 2024.

\bibitem[Zhang et~al.(2024{\natexlab{b}})Zhang, Zhang, Yao, Jia, Liu, Liu, and Liu]{zhang2024unlearncanvas}
Y.~Zhang, Y.~Zhang, Y.~Yao, J.~Jia, J.~Liu, X.~Liu, and S.~Liu, ``Unlearncanvas: A stylized image dataset to benchmark machine unlearning for diffusion models,'' \emph{arXiv preprint arXiv:2402.11846}, 2024.

\bibitem[Liu et~al.(2022{\natexlab{c}})Liu, Xu, Yuan, Wang, and Li]{liu2022right}
Y.~Liu, L.~Xu, X.~Yuan, C.~Wang, and B.~Li, ``The right to be forgotten in federated learning: An efficient realization with rapid retraining,'' in \emph{IEEE INFOCOM 2022-IEEE Conference on Computer Communications}.\hskip 1em plus 0.5em minus 0.4em\relax IEEE, 2022, pp. 1749--1758.

\bibitem[Halimi et~al.(2022)Halimi, Kadhe, Rawat, and Baracaldo]{halimi2022federated}
A.~Halimi, S.~Kadhe, A.~Rawat, and N.~Baracaldo, ``Federated unlearning: How to efficiently erase a client in fl?'' \emph{arXiv preprint arXiv:2207.05521}, 2022.

\bibitem[Jin et~al.(2023)Jin, Chen, Zhang, and Li]{jin2023forgettable}
R.~Jin, M.~Chen, Q.~Zhang, and X.~Li, ``Forgettable federated linear learning with certified data removal,'' \emph{arXiv preprint arXiv:2306.02216}, 2023.

\bibitem[Chen et~al.(2022)Chen, Zhang, Wang, Backes, Humbert, and Zhang]{chen2022graph}
M.~Chen, Z.~Zhang, T.~Wang, M.~Backes, M.~Humbert, and Y.~Zhang, ``Graph unlearning,'' in \emph{Proceedings of the 2022 ACM SIGSAC Conference on Computer and Communications Security}, 2022, pp. 499--513.

\bibitem[Chien et~al.(2022)Chien, Pan, and Milenkovic]{chien2022certified}
E.~Chien, C.~Pan, and O.~Milenkovic, ``Certified graph unlearning,'' \emph{arXiv preprint arXiv:2206.09140}, 2022.

\bibitem[Wu et~al.(2023{\natexlab{a}})Wu, Shen, Ning, Wang, and Wang]{wu2023certified}
K.~Wu, J.~Shen, Y.~Ning, T.~Wang, and W.~H. Wang, ``Certified edge unlearning for graph neural networks,'' in \emph{Proceedings of the 29th ACM SIGKDD Conference on Knowledge Discovery and Data Mining}, 2023, pp. 2606--2617.

\bibitem[Lu et~al.(2022)Lu, Welleck, Hessel, Jiang, Qin, West, Ammanabrolu, and Choi]{lu2022quark}
X.~Lu, S.~Welleck, J.~Hessel, L.~Jiang, L.~Qin, P.~West, P.~Ammanabrolu, and Y.~Choi, ``Quark: Controllable text generation with reinforced unlearning,'' \emph{Advances in neural information processing systems}, vol.~35, pp. 27\,591--27\,609, 2022.

\bibitem[Jang et~al.(2022)Jang, Yoon, Yang, Cha, Lee, Logeswaran, and Seo]{jang2022knowledge}
J.~Jang, D.~Yoon, S.~Yang, S.~Cha, M.~Lee, L.~Logeswaran, and M.~Seo, ``Knowledge unlearning for mitigating privacy risks in language models,'' \emph{arXiv preprint arXiv:2210.01504}, 2022.

\bibitem[Kumar et~al.(2022)Kumar, Gangadharaiah, and Roth]{kumar2022privacy}
V.~B. Kumar, R.~Gangadharaiah, and D.~Roth, ``Privacy adhering machine un-learning in nlp,'' \emph{arXiv preprint arXiv:2212.09573}, 2022.

\bibitem[Zhang et~al.(2023)Zhang, Wang, Xu, Wang, and Shi]{zhang2023forget}
E.~Zhang, K.~Wang, X.~Xu, Z.~Wang, and H.~Shi, ``Forget-me-not: Learning to forget in text-to-image diffusion models,'' \emph{arXiv preprint arXiv:2303.17591}, 2023.

\bibitem[Pawelczyk et~al.(2023)Pawelczyk, Neel, and Lakkaraju]{pawelczyk2023context}
M.~Pawelczyk, S.~Neel, and H.~Lakkaraju, ``In-context unlearning: Language models as few shot unlearners,'' \emph{arXiv preprint arXiv:2310.07579}, 2023.

\bibitem[Ishibashi and Shimodaira(2023)]{ishibashi2023knowledge}
Y.~Ishibashi and H.~Shimodaira, ``Knowledge sanitization of large language models,'' \emph{arXiv preprint arXiv:2309.11852}, 2023.

\bibitem[Wang et~al.(2024{\natexlab{a}})Wang, Wu, He, Chen, and McAuley]{wang2024large}
Y.~Wang, R.~Wu, Z.~He, X.~Chen, and J.~McAuley, ``Large scale knowledge washing,'' \emph{arXiv preprint arXiv:2405.16720}, 2024.

\bibitem[Liu et~al.(2024)Liu, Wang, Flanigan, and Liu]{liu2024large}
C.~Y. Liu, Y.~Wang, J.~Flanigan, and Y.~Liu, ``Large language model unlearning via embedding-corrupted prompts,'' \emph{arXiv preprint arXiv:2406.07933}, 2024.

\bibitem[Thaker et~al.(2024)Thaker, Maurya, and Smith]{thaker2024guardrail}
P.~Thaker, Y.~Maurya, and V.~Smith, ``Guardrail baselines for unlearning in llms,'' \emph{arXiv preprint arXiv:2403.03329}, 2024.

\bibitem[Kadhe et~al.(2024)Kadhe, Ahmed, Wei, Baracaldo, and Padhi]{kadhe2024split}
S.~R. Kadhe, F.~Ahmed, D.~Wei, N.~Baracaldo, and I.~Padhi, ``Split, unlearn, merge: Leveraging data attributes for more effective unlearning in llms,'' \emph{arXiv preprint arXiv:2406.11780}, 2024.

\bibitem[Gu et~al.(2024)Gu, Huang, Luo, Yao, Yang, Teng, and Wang]{gu2024meow}
T.~Gu, K.~Huang, R.~Luo, Y.~Yao, Y.~Yang, Y.~Teng, and Y.~Wang, ``Meow: Memory supervised llm unlearning via inverted facts,'' \emph{arXiv preprint arXiv:2409.11844}, 2024.

\bibitem[Wang et~al.(2024{\natexlab{b}})Wang, Zi, Sun, Zhao, and Qin]{wang2024rkld}
B.~Wang, Y.~Zi, Y.~Sun, Y.~Zhao, and B.~Qin, ``Rkld: Reverse kl-divergence-based knowledge distillation for unlearning personal information in large language models,'' \emph{arXiv preprint arXiv:2406.01983}, 2024.

\bibitem[Mekala et~al.(2024)Mekala, Dorna, Dubey, Lalwani, Koleczek, Rungta, Hasan, and Lobo]{mekala2024alternate}
A.~Mekala, V.~Dorna, S.~Dubey, A.~Lalwani, D.~Koleczek, M.~Rungta, S.~Hasan, and E.~Lobo, ``Alternate preference optimization for unlearning factual knowledge in large language models,'' \emph{arXiv preprint arXiv:2409.13474}, 2024.

\bibitem[Wu et~al.(2023{\natexlab{b}})Wu, Li, Xu, Dong, Wu, Bian, and Xiong]{wu2023depn}
X.~Wu, J.~Li, M.~Xu, W.~Dong, S.~Wu, C.~Bian, and D.~Xiong, ``Depn: Detecting and editing privacy neurons in pretrained language models,'' \emph{arXiv preprint arXiv:2310.20138}, 2023.

\bibitem[Barrett et~al.(2023)Barrett, Boyd, Bursztein, Carlini, Chen, Choi, Chowdhury, Christodorescu, Datta, Feizi, et~al.]{barrett2023identifying}
C.~Barrett, B.~Boyd, E.~Bursztein, N.~Carlini, B.~Chen, J.~Choi, A.~R. Chowdhury, M.~Christodorescu, A.~Datta, S.~Feizi \emph{et~al.}, ``Identifying and mitigating the security risks of generative ai,'' \emph{Foundations and Trends{\textregistered} in Privacy and Security}, vol.~6, no.~1, pp. 1--52, 2023.

\bibitem[Yu et~al.(2023)Yu, Jeoung, Kasi, Yu, and Ji]{yu2023unlearning}
C.~Yu, S.~Jeoung, A.~Kasi, P.~Yu, and H.~Ji, ``Unlearning bias in language models by partitioning gradients,'' in \emph{Findings of the Association for Computational Linguistics: ACL 2023}, 2023, pp. 6032--6048.

\bibitem[Ilharco et~al.(2022)Ilharco, Ribeiro, Wortsman, Gururangan, Schmidt, Hajishirzi, and Farhadi]{ilharco2022editing}
G.~Ilharco, M.~T. Ribeiro, M.~Wortsman, S.~Gururangan, L.~Schmidt, H.~Hajishirzi, and A.~Farhadi, ``Editing models with task arithmetic,'' \emph{arXiv preprint arXiv:2212.04089}, 2022.

\bibitem[Schwarzschild et~al.(2024)Schwarzschild, Feng, Maini, Lipton, and Kolter]{schwarzschild2024rethinking}
A.~Schwarzschild, Z.~Feng, P.~Maini, Z.~C. Lipton, and J.~Z. Kolter, ``Rethinking llm memorization through the lens of adversarial compression,'' \emph{arXiv preprint arXiv:2404.15146}, 2024.

\bibitem[Patil et~al.(2024)Patil, Hase, and Bansal]{patil2023can}
V.~Patil, P.~Hase, and M.~Bansal, ``Can sensitive information be deleted from llms? objectives for defending against extraction attacks,'' \emph{ICLR}, 2024.

\bibitem[Christiano et~al.(2017)Christiano, Leike, Brown, Martic, Legg, and Amodei]{christiano2017deep}
P.~F. Christiano, J.~Leike, T.~Brown, M.~Martic, S.~Legg, and D.~Amodei, ``Deep reinforcement learning from human preferences,'' \emph{Advances in neural information processing systems}, vol.~30, 2017.

\bibitem[Ziegler et~al.(2019)Ziegler, Stiennon, Wu, Brown, Radford, Amodei, Christiano, and Irving]{ziegler2019fine}
D.~M. Ziegler, N.~Stiennon, J.~Wu, T.~B. Brown, A.~Radford, D.~Amodei, P.~Christiano, and G.~Irving, ``Fine-tuning language models from human preferences,'' \emph{arXiv preprint arXiv:1909.08593}, 2019.

\bibitem[Ouyang et~al.(2022)Ouyang, Wu, Jiang, Almeida, Wainwright, Mishkin, Zhang, Agarwal, Slama, Ray, et~al.]{ouyang2022training}
L.~Ouyang, J.~Wu, X.~Jiang, D.~Almeida, C.~Wainwright, P.~Mishkin, C.~Zhang, S.~Agarwal, K.~Slama, A.~Ray \emph{et~al.}, ``Training language models to follow instructions with human feedback,'' \emph{Advances in neural information processing systems}, vol.~35, pp. 27\,730--27\,744, 2022.

\bibitem[Santacroce et~al.(2023)Santacroce, Lu, Yu, Li, and Shen]{santacroce2023efficient}
M.~Santacroce, Y.~Lu, H.~Yu, Y.~Li, and Y.~Shen, ``Efficient rlhf: Reducing the memory usage of ppo,'' \emph{arXiv preprint arXiv:2309.00754}, 2023.

\bibitem[Zheng et~al.(2023)Zheng, Dou, Gao, Hua, Shen, Wang, Liu, Jin, Liu, Zhou, et~al.]{zheng2023secrets}
R.~Zheng, S.~Dou, S.~Gao, Y.~Hua, W.~Shen, B.~Wang, Y.~Liu, S.~Jin, Q.~Liu, Y.~Zhou \emph{et~al.}, ``Secrets of rlhf in large language models part i: Ppo,'' \emph{arXiv preprint arXiv:2307.04964}, 2023.

\bibitem[Zhao et~al.(2023)Zhao, Joshi, Liu, Khalman, Saleh, and Liu]{zhao2023slic}
Y.~Zhao, R.~Joshi, T.~Liu, M.~Khalman, M.~Saleh, and P.~J. Liu, ``Slic-hf: Sequence likelihood calibration with human feedback,'' \emph{arXiv preprint arXiv:2305.10425}, 2023.

\bibitem[Azar et~al.(2024)Azar, Guo, Piot, Munos, Rowland, Valko, and Calandriello]{azar2024general}
M.~G. Azar, Z.~D. Guo, B.~Piot, R.~Munos, M.~Rowland, M.~Valko, and D.~Calandriello, ``A general theoretical paradigm to understand learning from human preferences,'' in \emph{International Conference on Artificial Intelligence and Statistics}.\hskip 1em plus 0.5em minus 0.4em\relax PMLR, 2024, pp. 4447--4455.

\bibitem[Hong et~al.(2024)Hong, Lee, and Thorne]{hong2024reference}
J.~Hong, N.~Lee, and J.~Thorne, ``Reference-free monolithic preference optimization with odds ratio,'' \emph{arXiv preprint arXiv:2403.07691}, 2024.

\bibitem[Ethayarajh et~al.(2024)Ethayarajh, Xu, Muennighoff, Jurafsky, and Kiela]{ethayarajh2024kto}
K.~Ethayarajh, W.~Xu, N.~Muennighoff, D.~Jurafsky, and D.~Kiela, ``Kto: Model alignment as prospect theoretic optimization,'' \emph{arXiv preprint arXiv:2402.01306}, 2024.

\bibitem[Yuan et~al.(2024)Yuan, Yuan, Tan, Wang, Huang, and Huang]{yuan2024rrhf}
H.~Yuan, Z.~Yuan, C.~Tan, W.~Wang, S.~Huang, and F.~Huang, ``Rrhf: Rank responses to align language models with human feedback,'' \emph{Advances in Neural Information Processing Systems}, vol.~36, 2024.

\bibitem[Radford et~al.(2019)Radford, Wu, Child, Luan, Amodei, Sutskever, et~al.]{radford2019language}
A.~Radford, J.~Wu, R.~Child, D.~Luan, D.~Amodei, I.~Sutskever \emph{et~al.}, ``Language models are unsupervised multitask learners,'' \emph{OpenAI blog}, vol.~1, no.~8, p.~9, 2019.

\bibitem[Loshchilov and Hutter(2017)]{loshchilov2017decoupled}
I.~Loshchilov and F.~Hutter, ``Decoupled weight decay regularization,'' \emph{arXiv preprint arXiv:1711.05101}, 2017.

\end{thebibliography}
\bibliographystyle{IEEEtranN}


\clearpage
\newpage 

\newpage
\section*{NeurIPS Paper Checklist}

\begin{enumerate}

\item {\bf Claims}
    \item[] Question: Do the main claims made in the abstract and introduction accurately reflect the paper's contributions and scope?
    \item[] Answer: \answerYes{} 
    \item[] Justification: Our main claims are outlined in the abstract and Section\,\ref{sec: intro}. Comprehensive details are provided in Sections\,\ref{sec: background},\,\ref{sec: NPO_limitations},\,\ref{sec: SimNPO} and \ref{sec: exps}.
    \item[] Guidelines:
    \begin{itemize}
        \item The answer NA means that the abstract and introduction do not include the claims made in the paper.
        \item The abstract and/or introduction should clearly state the claims made, including the contributions made in the paper and important assumptions and limitations. A No or NA answer to this question will not be perceived well by the reviewers. 
        \item The claims made should match theoretical and experimental results, and reflect how much the results can be expected to generalize to other settings. 
        \item It is fine to include aspirational goals as motivation as long as it is clear that these goals are not attained by the paper. 
    \end{itemize}

\item {\bf Limitations}
    \item[] Question: Does the paper discuss the limitations of the work performed by the authors?
    \item[] Answer: \answerYes{} 
    \item[] Justification: Limitations of our work are discussed in Appendix\,\ref{appx: limitations}.
    \item[] Guidelines:
    \begin{itemize}
        \item The answer NA means that the paper has no limitation while the answer No means that the paper has limitations, but those are not discussed in the paper. 
        \item The authors are encouraged to create a separate "Limitations" section in their paper.
        \item The paper should point out any strong assumptions and how robust the results are to violations of these assumptions (e.g., independence assumptions, noiseless settings, model well-specification, asymptotic approximations only holding locally). The authors should reflect on how these assumptions might be violated in practice and what the implications would be.
        \item The authors should reflect on the scope of the claims made, e.g., if the approach was only tested on a few datasets or with a few runs. In general, empirical results often depend on implicit assumptions, which should be articulated.
        \item The authors should reflect on the factors that influence the performance of the approach. For example, a facial recognition algorithm may perform poorly when image resolution is low or images are taken in low lighting. Or a speech-to-text system might not be used reliably to provide closed captions for online lectures because it fails to handle technical jargon.
        \item The authors should discuss the computational efficiency of the proposed algorithms and how they scale with dataset size.
        \item If applicable, the authors should discuss possible limitations of their approach to address problems of privacy and fairness.
        \item While the authors might fear that complete honesty about limitations might be used by reviewers as grounds for rejection, a worse outcome might be that reviewers discover limitations that aren't acknowledged in the paper. The authors should use their best judgment and recognize that individual actions in favor of transparency play an important role in developing norms that preserve the integrity of the community. Reviewers will be specifically instructed to not penalize honesty concerning limitations.
    \end{itemize}

\item {\bf Theory assumptions and proofs}
    \item[] Question: For each theoretical result, does the paper provide the full set of assumptions and a complete (and correct) proof?
    \item[] Answer: \answerYes{} 
    \item[] Justification: Theoretical analyses of the gradient and unlearning speed of our method are provided in Appendix\,\ref{app: grad_deri} and \ref{app: explanation_divergence}.
    \item[] Guidelines:
    \begin{itemize}
        \item The answer NA means that the paper does not include theoretical results. 
        \item All the theorems, formulas, and proofs in the paper should be numbered and cross-referenced.
        \item All assumptions should be clearly stated or referenced in the statement of any theorems.
        \item The proofs can either appear in the main paper or the supplemental material, but if they appear in the supplemental material, the authors are encouraged to provide a short proof sketch to provide intuition. 
        \item Inversely, any informal proof provided in the core of the paper should be complemented by formal proofs provided in appendix or supplemental material.
        \item Theorems and Lemmas that the proof relies upon should be properly referenced. 
    \end{itemize}

    \item {\bf Experimental result reproducibility}
    \item[] Question: Does the paper fully disclose all the information needed to reproduce the main experimental results of the paper to the extent that it affects the main claims and/or conclusions of the paper (regardless of whether the code and data are provided or not)?
    \item[] Answer: \answerYes{} 
    \item[] Justification: The experimental setup is detailed in Section\,\ref{sec: exps} and Appendix\,\ref{app: exp_setup}.
    \item[] Guidelines:
    \begin{itemize}
        \item The answer NA means that the paper does not include experiments.
        \item If the paper includes experiments, a No answer to this question will not be perceived well by the reviewers: Making the paper reproducible is important, regardless of whether the code and data are provided or not.
        \item If the contribution is a dataset and/or model, the authors should describe the steps taken to make their results reproducible or verifiable. 
        \item Depending on the contribution, reproducibility can be accomplished in various ways. For example, if the contribution is a novel architecture, describing the architecture fully might suffice, or if the contribution is a specific model and empirical evaluation, it may be necessary to either make it possible for others to replicate the model with the same dataset, or provide access to the model. In general. releasing code and data is often one good way to accomplish this, but reproducibility can also be provided via detailed instructions for how to replicate the results, access to a hosted model (e.g., in the case of a large language model), releasing of a model checkpoint, or other means that are appropriate to the research performed.
        \item While NeurIPS does not require releasing code, the conference does require all submissions to provide some reasonable avenue for reproducibility, which may depend on the nature of the contribution. For example
        \begin{enumerate}
            \item If the contribution is primarily a new algorithm, the paper should make it clear how to reproduce that algorithm.
            \item If the contribution is primarily a new model architecture, the paper should describe the architecture clearly and fully.
            \item If the contribution is a new model (e.g., a large language model), then there should either be a way to access this model for reproducing the results or a way to reproduce the model (e.g., with an open-source dataset or instructions for how to construct the dataset).
            \item We recognize that reproducibility may be tricky in some cases, in which case authors are welcome to describe the particular way they provide for reproducibility. In the case of closed-source models, it may be that access to the model is limited in some way (e.g., to registered users), but it should be possible for other researchers to have some path to reproducing or verifying the results.
        \end{enumerate}
    \end{itemize}

\item {\bf Open access to data and code}
    \item[] Question: Does the paper provide open access to the data and code, with sufficient instructions to faithfully reproduce the main experimental results, as described in supplemental material?
    \item[] Answer: \answerYes{} 
    \item[] Justification: The implementation code is included in the supplementary materials.
    \item[] Guidelines:
    \begin{itemize}
        \item The answer NA means that paper does not include experiments requiring code.
        \item Please see the NeurIPS code and data submission guidelines (\url{https://nips.cc/public/guides/CodeSubmissionPolicy}) for more details.
        \item While we encourage the release of code and data, we understand that this might not be possible, so “No” is an acceptable answer. Papers cannot be rejected simply for not including code, unless this is central to the contribution (e.g., for a new open-source benchmark).
        \item The instructions should contain the exact command and environment needed to run to reproduce the results. See the NeurIPS code and data submission guidelines (\url{https://nips.cc/public/guides/CodeSubmissionPolicy}) for more details.
        \item The authors should provide instructions on data access and preparation, including how to access the raw data, preprocessed data, intermediate data, and generated data, etc.
        \item The authors should provide scripts to reproduce all experimental results for the new proposed method and baselines. If only a subset of experiments are reproducible, they should state which ones are omitted from the script and why.
        \item At submission time, to preserve anonymity, the authors should release anonymized versions (if applicable).
        \item Providing as much information as possible in supplemental material (appended to the paper) is recommended, but including URLs to data and code is permitted.
    \end{itemize}

\item {\bf Experimental setting/details}
    \item[] Question: Does the paper specify all the training and test details (e.g., data splits, hyperparameters, how they were chosen, type of optimizer, etc.) necessary to understand the results?
    \item[] Answer: \answerYes{} 
    \item[] Justification: The experimental setup is detailed in Section\,\ref{sec: exps} and Appendix\,\ref{app: exp_setup}.
    \item[] Guidelines:
    \begin{itemize}
        \item The answer NA means that the paper does not include experiments.
        \item The experimental setting should be presented in the core of the paper to a level of detail that is necessary to appreciate the results and make sense of them.
        \item The full details can be provided either with the code, in appendix, or as supplemental material.
    \end{itemize}

\item {\bf Experiment statistical significance}
    \item[] Question: Does the paper report error bars suitably and correctly defined or other appropriate information about the statistical significance of the experiments?
    \item[] Answer: \answerYes{} 
    \item[] Justification: To ensure a fair evaluation, we perform multiple runs for each setting and report the average performance. As an example, Fig.\,\ref{fig: relearn} presents the results with error bars.

    \item[] Guidelines:
    \begin{itemize}
        \item The answer NA means that the paper does not include experiments.
        \item The authors should answer "Yes" if the results are accompanied by error bars, confidence intervals, or statistical significance tests, at least for the experiments that support the main claims of the paper.
        \item The factors of variability that the error bars are capturing should be clearly stated (for example, train/test split, initialization, random drawing of some parameter, or overall run with given experimental conditions).
        \item The method for calculating the error bars should be explained (closed form formula, call to a library function, bootstrap, etc.)
        \item The assumptions made should be given (e.g., Normally distributed errors).
        \item It should be clear whether the error bar is the standard deviation or the standard error of the mean.
        \item It is OK to report 1-sigma error bars, but one should state it. The authors should preferably report a 2-sigma error bar than state that they have a 96\% CI, if the hypothesis of Normality of errors is not verified.
        \item For asymmetric distributions, the authors should be careful not to show in tables or figures symmetric error bars that would yield results that are out of range (e.g. negative error rates).
        \item If error bars are reported in tables or plots, The authors should explain in the text how they were calculated and reference the corresponding figures or tables in the text.
    \end{itemize}

\item {\bf Experiments compute resources}
    \item[] Question: For each experiment, does the paper provide sufficient information on the computer resources (type of compute workers, memory, time of execution) needed to reproduce the experiments?
    \item[] Answer: \answerYes{} 
    \item[] Justification: The computer resources are detailed in Appendix\,\ref{app: exp_setup}.
    \item[] Guidelines:
    \begin{itemize}
        \item The answer NA means that the paper does not include experiments.
        \item The paper should indicate the type of compute workers CPU or GPU, internal cluster, or cloud provider, including relevant memory and storage.
        \item The paper should provide the amount of compute required for each of the individual experimental runs as well as estimate the total compute. 
        \item The paper should disclose whether the full research project required more compute than the experiments reported in the paper (e.g., preliminary or failed experiments that didn't make it into the paper). 
    \end{itemize}
    
\item {\bf Code of ethics}
    \item[] Question: Does the research conducted in the paper conform, in every respect, with the NeurIPS Code of Ethics \url{https://neurips.cc/public/EthicsGuidelines}?
    \item[] Answer: \answerYes{} 
    \item[] Justification: We have taken all necessary steps to ensure author anonymity.
    \item[] Guidelines:
    \begin{itemize}
        \item The answer NA means that the authors have not reviewed the NeurIPS Code of Ethics.
        \item If the authors answer No, they should explain the special circumstances that require a deviation from the Code of Ethics.
        \item The authors should make sure to preserve anonymity (e.g., if there is a special consideration due to laws or regulations in their jurisdiction).
    \end{itemize}

\item {\bf Broader impacts}
    \item[] Question: Does the paper discuss both potential positive societal impacts and negative societal impacts of the work performed?
    \item[] Answer: \answerYes{} 
    \item[] Justification: Impacts of our work are discussed in Appendix\,\ref{appx: impact}.
    \item[] Guidelines:
    \begin{itemize}
        \item The answer NA means that there is no societal impact of the work performed.
        \item If the authors answer NA or No, they should explain why their work has no societal impact or why the paper does not address societal impact.
        \item Examples of negative societal impacts include potential malicious or unintended uses (e.g., disinformation, generating fake profiles, surveillance), fairness considerations (e.g., deployment of technologies that could make decisions that unfairly impact specific groups), privacy considerations, and security considerations.
        \item The conference expects that many papers will be foundational research and not tied to particular applications, let alone deployments. However, if there is a direct path to any negative applications, the authors should point it out. For example, it is legitimate to point out that an improvement in the quality of generative models could be used to generate deepfakes for disinformation. On the other hand, it is not needed to point out that a generic algorithm for optimizing neural networks could enable people to train models that generate Deepfakes faster.
        \item The authors should consider possible harms that could arise when the technology is being used as intended and functioning correctly, harms that could arise when the technology is being used as intended but gives incorrect results, and harms following from (intentional or unintentional) misuse of the technology.
        \item If there are negative societal impacts, the authors could also discuss possible mitigation strategies (e.g., gated release of models, providing defenses in addition to attacks, mechanisms for monitoring misuse, mechanisms to monitor how a system learns from feedback over time, improving the efficiency and accessibility of ML).
    \end{itemize}
    
\item {\bf Safeguards}
    \item[] Question: Does the paper describe safeguards that have been put in place for responsible release of data or models that have a high risk for misuse (e.g., pretrained language models, image generators, or scraped datasets)?
    \item[] Answer: \answerNA{} 
    \item[] Justification: Built upon existing open-source models, our method is intended solely for unlearning benchmark evaluations and does not incorporate any data or models with a high potential for misuse.

    \item[] Guidelines:
    \begin{itemize}
        \item The answer NA means that the paper poses no such risks.
        \item Released models that have a high risk for misuse or dual-use should be released with necessary safeguards to allow for controlled use of the model, for example by requiring that users adhere to usage guidelines or restrictions to access the model or implementing safety filters. 
        \item Datasets that have been scraped from the Internet could pose safety risks. The authors should describe how they avoided releasing unsafe images.
        \item We recognize that providing effective safeguards is challenging, and many papers do not require this, but we encourage authors to take this into account and make a best faith effort.
    \end{itemize}

\item {\bf Licenses for existing assets}
    \item[] Question: Are the creators or original owners of assets (e.g., code, data, models), used in the paper, properly credited and are the license and terms of use explicitly mentioned and properly respected?
    \item[] Answer: \answerYes{} 
    \item[] Justification: We acknowledge and cite the original sources of the codebase and dataset employed in this study.
    
    \item[] Guidelines:
    \begin{itemize}
        \item The answer NA means that the paper does not use existing assets.
        \item The authors should cite the original paper that produced the code package or dataset.
        \item The authors should state which version of the asset is used and, if possible, include a URL.
        \item The name of the license (e.g., CC-BY 4.0) should be included for each asset.
        \item For scraped data from a particular source (e.g., website), the copyright and terms of service of that source should be provided.
        \item If assets are released, the license, copyright information, and terms of use in the package should be provided. For popular datasets, \url{paperswithcode.com/datasets} has curated licenses for some datasets. Their licensing guide can help determine the license of a dataset.
        \item For existing datasets that are re-packaged, both the original license and the license of the derived asset (if it has changed) should be provided.
        \item If this information is not available online, the authors are encouraged to reach out to the asset's creators.
    \end{itemize}

\item {\bf New assets}
    \item[] Question: Are new assets introduced in the paper well documented and is the documentation provided alongside the assets?
    \item[] Answer: \answerYes{} 
    \item[] Justification: The code used to conduct our experiments is included in the supplementary material.
    \item[] Guidelines:
    \begin{itemize}
        \item The answer NA means that the paper does not release new assets.
        \item Researchers should communicate the details of the dataset/code/model as part of their submissions via structured templates. This includes details about training, license, limitations, etc. 
        \item The paper should discuss whether and how consent was obtained from people whose asset is used.
        \item At submission time, remember to anonymize your assets (if applicable). You can either create an anonymized URL or include an anonymized zip file.
    \end{itemize}

\item {\bf Crowdsourcing and research with human subjects}
    \item[] Question: For crowdsourcing experiments and research with human subjects, does the paper include the full text of instructions given to participants and screenshots, if applicable, as well as details about compensation (if any)? 
    \item[] Answer: \answerNA{} 
    \item[] Justification: This work does not involve any human subjects.
    \item[] Guidelines:
    \begin{itemize}
        \item The answer NA means that the paper does not involve crowdsourcing nor research with human subjects.
        \item Including this information in the supplemental material is fine, but if the main contribution of the paper involves human subjects, then as much detail as possible should be included in the main paper. 
        \item According to the NeurIPS Code of Ethics, workers involved in data collection, curation, or other labor should be paid at least the minimum wage in the country of the data collector. 
    \end{itemize}

\item {\bf Institutional review board (IRB) approvals or equivalent for research with human subjects}
    \item[] Question: Does the paper describe potential risks incurred by study participants, whether such risks were disclosed to the subjects, and whether Institutional Review Board (IRB) approvals (or an equivalent approval/review based on the requirements of your country or institution) were obtained?
    \item[] Answer: \answerNA{} 
    \item[] Justification: This work does not involve any human subjects.
    \item[] Guidelines:
    \begin{itemize}
        \item The answer NA means that the paper does not involve crowdsourcing nor research with human subjects.
        \item Depending on the country in which research is conducted, IRB approval (or equivalent) may be required for any human subjects research. If you obtained IRB approval, you should clearly state this in the paper. 
        \item We recognize that the procedures for this may vary significantly between institutions and locations, and we expect authors to adhere to the NeurIPS Code of Ethics and the guidelines for their institution. 
        \item For initial submissions, do not include any information that would break anonymity (if applicable), such as the institution conducting the review.
    \end{itemize}

\item {\bf Declaration of LLM usage}
    \item[] Question: Does the paper describe the usage of LLMs if it is an important, original, or non-standard component of the core methods in this research? Note that if the LLM is used only for writing, editing, or formatting purposes and does not impact the core methodology, scientific rigorousness, or originality of the research, declaration is not required.
    \item[] Answer: \answerNA{} 
    \item[] Justification: We use LLMs solely for writing refinement.
    \item[] Guidelines:
    \begin{itemize}
        \item The answer NA means that the core method development in this research does not involve LLMs as any important, original, or non-standard components.
        \item Please refer to our LLM policy (\url{https://neurips.cc/Conferences/2025/LLM}) for what should or should not be described.
    \end{itemize}

\end{enumerate}

\appendix

\clearpage
\newpage
\onecolumn
\section*{\Large{Appendix}}

\setcounter{section}{0}
\setcounter{figure}{0}
\makeatletter 
\renewcommand{\thefigure}{A\arabic{figure}}
\renewcommand{\theHfigure}{A\arabic{figure}}
\renewcommand{\thetable}{A\arabic{table}}
\renewcommand{\theHtable}{A\arabic{table}}

\makeatother
\setcounter{table}{0}

\setcounter{mylemma}{0}
\renewcommand{\themylemma}{A\arabic{mylemma}}
\setcounter{equation}{0}
\renewcommand{\theequation}{A\arabic{equation}}

\section{A Summary of the Unlearning Tasks and Evaluation Metrics}
\label{app: evaluation}

\begin{table*}[htbp]
\centering
\vspace{-3mm}
\caption{\footnotesize{Summary of unlearning efficacy and utility metrics across different unlearning benchmarks. The arrows indicate the directions for better performance (↑ for higher values, ↓ for lower values, $\rightarrow 0$ for closer to 0). 
}}
\resizebox{\textwidth}{!}{
\begin{tabular}{c|c|c|cc|cc}
\toprule
\midrule
\multirow{1}{*}{\textbf{Benchmark}} & \multirow{1}{*}{\textbf{LLM to be used}} & \multirow{1}{*}{\textbf{Task Description}} & \multicolumn{2}{c|}{\textbf{Unlearning Effectiveness}} & \multicolumn{2}{c}{\textbf{Utility Preservation}} \\ 
\midrule

\multirow{3}{*}{TOFU} & \multirow{3}{*}{LLaMA-2-chat 7B} & \multirow{3}{*}{\parbox{5cm}{\centering Unlearning fictitious authors from a synthetic Q\&A dataset}} & 
\multicolumn{1}{c}{
\begin{tabular}[c]{@{}c@{}}
     {Forget quality} (measured by\\
      truth ratios of forget samples)
    \end{tabular}
} & \multicolumn{1}{c|}{↑} & \multicolumn{1}{c}{
\begin{tabular}[c]{@{}c@{}}
     {Model utility}\\
     ( harmonic mean of 9 utility metrics)
    \end{tabular}
    }
 & \multicolumn{1}{c}{↑} \\ 

& & & \multicolumn{1}{c}{Probability on $\mathcal{D}_f$} & \multicolumn{1}{c|}{↓} & \multicolumn{1}{c}{Probability on $\mathcal{D}_r$/$\mathcal{D}_{\text{real\_author}}$/$\mathcal{D}_{\text{world\_facts}}$} & \multicolumn{1}{c}{↑} \\
& & & \multicolumn{1}{c}{Rouge-L on $\mathcal{D}_f$} & \multicolumn{1}{c|}{↓} & \multicolumn{1}{c}{Rouge-L on $\mathcal{D}_r$/$\mathcal{D}_{\text{real\_author}}$/$\mathcal{D}_{\text{world\_facts}}$} & \multicolumn{1}{c}{↑} \\
& & & \multicolumn{1}{c}{Truth ratio on $\mathcal{D}_f$} & \multicolumn{1}{c|}{↑} & \multicolumn{1}{c}{Truth ratio on $\mathcal{D}_r$/$\mathcal{D}_{\text{real\_author}}$/$\mathcal{D}_{\text{world\_facts}}$} & \multicolumn{1}{c}{↑} \\
\midrule

\multirow{3}{*}{MUSE} & \multirow{3}{*}{\parbox{2cm}{\centering ICLM-7B \\ LLaMA-2 7B }} & 
\multirow{3}{*}{\parbox{5cm}{\centering Unlearning real-world knowledge from texts about Harry Potter and BBC News}} & \multicolumn{1}{c}{KnowMem on $\mathcal{D}_f$} & \multicolumn{1}{c|}{↓} & \multicolumn{1}{c}{} & \multicolumn{1}{c}{} \\
& & & \multicolumn{1}{c}{VerbMem on $\mathcal{D}_f$} & \multicolumn{1}{c|}{↓} & \multicolumn{1}{c}{KnowMem on $\mathcal{D}_r$} & \multicolumn{1}{c}{↑} \\
& & & \multicolumn{1}{c}{PrivLeak} & \multicolumn{1}{c|}{$\rightarrow$ 0} & \multicolumn{1}{c}{} & \multicolumn{1}{c}{} \\
\midrule

\multirow{2}{*}{WMDP} & \multirow{2}{*}{Zephyr-7B-beta} & \multirow{2}{*}{\parbox{5cm}{\centering Unlearning hazardous knowledge from biosecurity texts}} & 
\multirow{2}{*}{Accuracy on WMDP-Bio} & \multirow{2}{*}{↓} & \multirow{2}{*}{Accuracy on MMLU} & \multirow{2}{*}{↑} \\
& & &  &  &  & \multicolumn{1}{c}{} \\

\midrule
\bottomrule
\end{tabular}
}
\label{tab:tasks_evaluations}
\end{table*}
\section{Additional on the sensitivity of NPO to reference model}
\label{app: npo_reference_model}

To examine the sensitivity of NPO to its reference model choice $\bigl(\btheta_{\mathrm{ref}}$, used interchangeably with $\pi_{\mathrm{ref}}\bigr)$, we design a perturbed reference model $\btheta_{\mathrm{ref}}^\prime$ by averaging $\btheta_{\mathrm{ref}}$ with a randomly initialized model.  
We then apply NPO with $\btheta_{\mathrm{ref}}^\prime$ as the reference on the TOFU Forget05, following the same experimental setup as in Fig.\,\ref{fig: intro_fig} (c).  
This perturbation leads to a dramatic drop in forget quality—from $0.79$ with $\btheta_{\mathrm{ref}}$ to $0.27$ with $\btheta_{\mathrm{ref}}^\prime$—while the model utility remains largely unaffected ($0.57$ vs.\ $0.52$).  
These results highlight the crucial role of the reference model in ensuring reliable unlearning performance.

\begin{figure}[htb]
\vspace{-3mm}
\centerline{
\hspace*{0mm}\includegraphics[width=.26\textwidth,height=!]{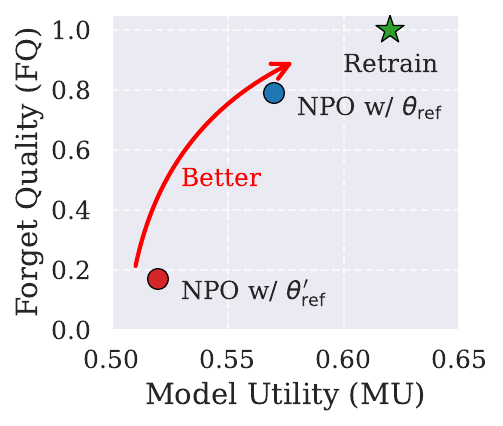}
}
\caption{
\small{Forget quality and model utility of NPO w/ $\btheta_{\mathrm{ref}}^\prime$, NPO w/ $\btheta_{\mathrm{ref}}$ and Retrain on TOFU Forget05. The figure format follows Fig.\,\ref{fig: intro_fig} (c).}
}
  \label{fig: npo_reference_model}
  \vspace*{-3mm}
\end{figure}
\section{Additional Setup and Results on Unlearning vs. Data Memorization}
\label{app: strongly_weakly_memorized}

\begin{wraptable}{r}{0.35\textwidth}
\centering
\vspace{-6mm}
\caption{\small{Unlearning performance on differently memorized forget sets  $\mathcal{D}_{\mathrm{f},1}$ and $\mathcal{D}_{\mathrm{f},2}$  
in TOFU.}}
\vspace{1mm}
\label{tab: strongly_weakly_memorized}
\resizebox{0.35\textwidth}{!}{
\begin{tabular}{c|cc|c}
\toprule[1pt]
\midrule
           & FQ on $\mathcal{D}_{\mathrm{f},1}$ & FQ on $\mathcal{D}_{\mathrm{f},2}$ & Utility \\
\midrule
Original   & 0.00                                  & 0.01                                  & 0.62    \\
\midrule
NPO        & 0.00                                  & 0.47                               & 0.49    \\
\rowcolor{Gray}
SimNPO     & 0.70                                  & 0.70                                  & 0.57    \\
\midrule
\bottomrule[1pt]
\end{tabular}
}
\vspace*{-3mm}
\end{wraptable}

We use TOFU Forget05 as the forget set \({\mathcal{D}_{\mathrm{f}}}\), splitting it evenly into $\mathcal{D}_{\mathrm{f},1}$ and $\mathcal{D}_{\mathrm{f},2}$. The divided subsets $\mathcal{D}_{\mathrm{f},1}$ and $\mathcal{D}_{\mathrm{f},2}$ follow the same distribution of fictitious author information. We fine-tune the LLaMA-2 7B chat model on the original retain set of TOFU together with $\mathcal{D}_{\mathrm{f},1}$, \textit{i.e.}, \({\mathcal{D}_{\text{retain}} \cup \mathcal{D}_{\mathrm{f},1}}\), to obtain   the original model before unlearning. The resulting original model strongly memorizes $\mathcal{D}_{\mathrm{f},1}$ but least memorizes $\mathcal{D}_{\mathrm{f},2}$, despite both being drawn from the same distribution. We then perform unlearning using SimNPO and NPO over \({\mathcal{D}_{\mathrm{f},1} \cup \mathcal{D}_{\mathrm{f},2}}\). The unlearning performance, measured in terms of forget quality (FQ) and model utility, is presented in Table~\ref{tab: strongly_weakly_memorized}

As shown in Table~\ref{tab: strongly_weakly_memorized}, since the original model was trained on $\mathcal{D}_{\mathrm{f},1}$, its prediction loss \(-\log (\pi_{\text{ref}})\) on $\mathcal{D}_{\mathrm{f},1}$ is relatively small, leading to a higher prediction probability \(\pi_{\text{ref}}\) on $\mathcal{D}_{\mathrm{f},1}$. Consequently, the NPO gradient smoothing term in 
\eqref{eq:NPO_grads}
becomes relatively smaller for $\mathcal{D}_{\mathrm{f},1}$ due to the reference model's bias \(\pi_{\text{ref}}\) on $\mathcal{D}_{\mathrm{f},1}$. As a result, NPO allocates less first-order optimization power to $\mathcal{D}_{\mathrm{f},1}$ and focuses more on $\mathcal{D}_{\mathrm{f},2}$. This prevents NPO from effectively forgetting $\mathcal{D}_{\mathrm{f},1}$, potentially causing under-unlearning and ultimately reducing the FQ of $\mathcal{D}_{\mathrm{f},1}$ to nearly zero. In contrast, SimNPO, by leveraging a reference-model-free reward, achieves a much smaller FQ difference between $\mathcal{D}_{\mathrm{f},1}$ and $\mathcal{D}_{\mathrm{f},2}$ while delivering higher FQ for both datasets compared to NPO. Furthermore, SimNPO demonstrates better model utility relative to NPO.
\section{Ablation Studies on SimNPO's Hyperparameter Selection}
\label{app: ablation_hyper}

As shown in \eqref{eq:SimNPO_loss},
$\beta$ and $\gamma$ are the two  hyperparameters that control the unlearning effectiveness and utility preservation of SimNPO. Similar to NPO, $\beta$ is a temperature hyperparameter used to regulate the intensity of unlearning but normalized by the response length $|y|$ in SimNPO.  As $\beta \to 0$, SimNPO approaches weighted GA in Fig.\,\ref{fig: illustration_length_normalization}.  $\gamma$ is the reward margin parameter from SimPO, which introduces a constant shift to the (per-sample) prediction loss $-(\beta/|y|)\log \pi_{\btheta}(y|x)$ in SimNPO. Consequently, a larger $\gamma$ imposes a stricter unlearning margin, which could further suppress the model utility. 

\begin{figure}[htb]
\center
\begin{tabular}{cc}
\hspace*{-3mm}
\includegraphics[width=0.245\textwidth,height=!]{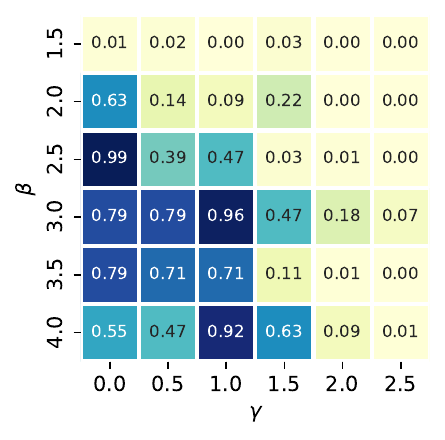} 
&
\hspace*{-6mm}
\includegraphics[width=0.245\textwidth,height=!]{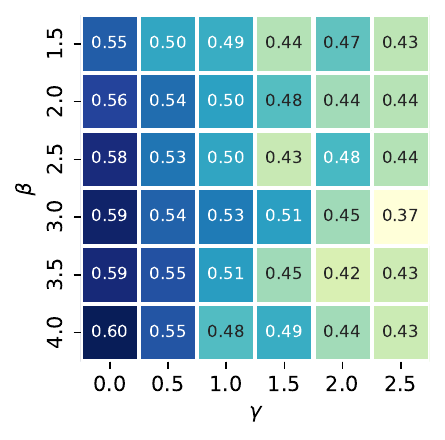}
\\
\hspace*{2mm} \scriptsize{(a) Forget quality} & \scriptsize{(b) Model utility}
\end{tabular}
\caption{\small{Forget quality (a) and model utility (b) of SimNPO under different combinations of $\beta$ and $\gamma$ on TOFU Forget05.}
}
  \label{fig: hyper_parameter_SimNPO}
\end{figure}

\textbf{Fig.\,\ref{fig: hyper_parameter_SimNPO}-(a)} and \textbf{Fig.\,\ref{fig: hyper_parameter_SimNPO}-(b)}  illustrate the forget quality and model utility of SimNPO under various values of $\beta$ and $\gamma$ on TOFU forget05. The results show that when $\beta$ is too small or $\gamma$ is too large, forget quality tends to decrease towards zero. Additionally, for a fixed $\beta$, increasing $\gamma$ leads to lower model utility. Notably, setting $\gamma = 0$ consistently yields the best balance between unlearning performance and utility preservation across different $\beta$ values, which supports our choice of $\gamma = 0$ in SimNPO.
\section{Gradient Analysis of SimNPO}
\label{app: grad_deri}

Following is the detailed derivation of \eqref{eq: grad_SimNPO_Nogamma}. First, let $\mathrm{R} = \frac{\log \pi_{\btheta} (y | x) + \gamma |y| / \beta}{|y|}$. We then have the following steps:
\begin{align}
\nabla_{\btheta} \ell_{\mathrm{SimNPO}}(\boldsymbol{\theta}) & = \mathbb E_{(x,y) \in \Df}  \nabla_{\btheta} \left[-\frac{2}{\beta} \log\sigma(-\beta \mathrm{R})\right] \\
& = \mathbb E_{(x,y) \in \Df} \nabla_{\btheta}\left[\frac{2}{\beta}\log\sigma(1+ \exp(\beta \mathrm{R}))\right] \\
& = \mathbb E_{(x,y) \in \Df} \left[\frac{2}{\beta} \cdot \frac{\beta\exp(\beta \mathrm{R})}{1+\exp(\beta \mathrm{R})}\cdot\nabla_{\btheta}\mathrm{R}\right] \\
& = \mathbb E_{(x,y) \in \Df} \left[\frac{2\exp(\beta \frac{\log  \pi_{\btheta} (y | x)+\gamma|y|/\beta}{|y|})}{1+\exp(\beta \frac{\log  \pi_{\btheta} (y | x)+\gamma|y|/\beta}{|y|})}\cdot\frac{1}{|y|}\cdot\nabla_{\btheta}{\log  \pi_{\btheta} (y | x)}\right]
    \label{eq:SimNPO_grads_full}
\end{align}
When $\gamma = 0$, the gradient simplifies to the following, which matches \eqref{eq: grad_SimNPO_Nogamma}:
\begin{align}
\nabla_{\btheta} \ell_{\mathrm{SimNPO}}(\boldsymbol{\theta}) & = \mathbb E_{(x,y) \in \Df} \left[\frac{2\exp(\frac{\beta\log  \pi_{\btheta} (y | x)}{|y|})}{1+\exp( \frac{\beta\log  \pi_{\btheta} (y | x)}{|y|})}\cdot\frac{1}{|y|}\cdot\nabla_{\btheta}{\log  \pi_{\btheta} (y | x)}\right] \\
&= \mathbb E_{(x,y) \in \Df} \left [ 
\frac{
2 (\pi_{\btheta}(y|x))^{\beta/|y|} 
}{1+ (\pi_{\btheta}(y|x))^{\beta/|y|}   }
\cdot
\frac{1}{|y|}
\cdot
\nabla_\btheta \log\pi_{\btheta}(y|x)
\right ]
    \label{eq:SimNPO_grads_gamma_0}
\end{align}
\section{Further Results on Response Length Normalization in SimNPO}
\label{app: length_normalization}

\begin{figure}[htb]
\centerline{
\hspace*{0mm}\includegraphics[width=.26\textwidth,height=!]{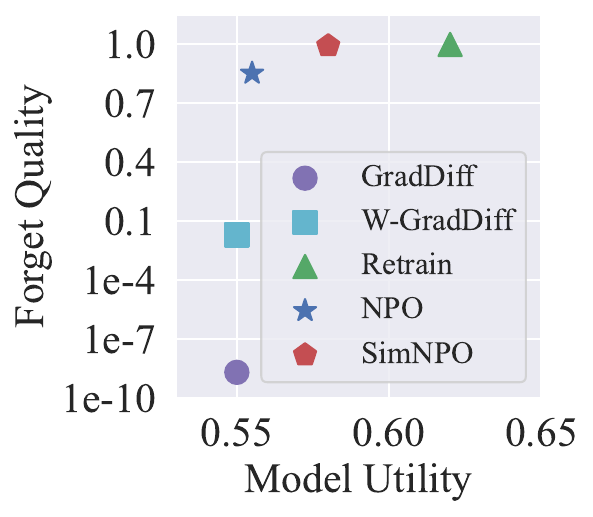}
}
\caption{
\small{Forget quality vs. model utility on TOFU Forget05. Weighted-GradDiff (W-GradDiff) is SimNPO at $\beta = 0$.}
}
  \label{fig: illustration_length_normalization}
\end{figure}

To better illustrate the role of length-normalization, we consider an extreme case: when $\beta \to 0$, the gradient of SimNPO degenerates into length-normalization weighted-GradDiff, while the gradient of NPO degenerates into GradDiff. In \textbf{Fig.\,\ref{fig: illustration_length_normalization}}-(a), we further compare the effects of weighted-GradDiff, GradDiff, NPO, and SimNPO. It can be observed that, due to the impact of length-normalization, the forget quality of weighted GradDiff is significantly better than that of GradDiff. This observation also explains why SimNPO achieves better forget quality compared to NPO.

\section{Further Analyses on Unlearning Speed}
\label{app: explanation_divergence}

The term ``unlearning speed'' or ``'divergence rate' refers to the optimization divergence from the pre-trained state, describing the process of deviating from the converged pre-trained model state to reverse the existing learning of the forgotten data. We present some further analyses for the unlearning speed of NPO and SimNPO.
Define $\log\widebar{\pi}_{\btheta}(y | x )= \log\pi_{\btheta}(y | x)/|y|$. Reorganizing the NPO gradient  formula in~\eqref{eq:NPO_grads}, 
and ignoring the reference model (or when $\pi_{\mathrm{ref}}(y|x)\approx1$), we have
\begin{align}
   \nabla_{\btheta} \ell_{\mathrm{NPO}}(\boldsymbol{\theta}) 
&=
   \mathbb \E_{(x,y) \in \Df}  \left[ \underbrace{
\left ( \frac{2
\widebar{\pi}_{\btheta}(y | x )^{|y|\beta}
}
{ \widebar{\pi}_{\btheta}(y | x )^{|y|\beta} + 1 }  \right)|y| }_{w(x,y)}
\cdot
{\nabla_\btheta \log\widebar{\pi}_{\btheta}(y | x)}
\right ].\notag
\end{align}
Suppose $ \log\widebar{\pi}_{\btheta}(y | x)$ is linear in $\btheta$ and the normalized gradient $ \nabla_\btheta\log\widebar{\pi}_{\btheta}(y |x)=\bigO(1).$ Then loosely speaking, the NPO dynamics satisfies the equation $\nabla_t\btheta(t)\approx -2|y|\cdot\exp(\beta|y|\btheta(t))$. Assuming  $\btheta(0)=\mathbf{0}$ and $\beta\ll1$,  this yields the solution $\btheta(t)=-\frac{1}{\beta|y|}\log(1+2\beta|y|^2 t)$, suggesting that the models uses $\bigO(\frac{(1/\eps)^{\beta|y|}-1}{\beta|y|^2\eta})=\bigO(\frac{\log(1/\eps)}{|y|\eta})$ steps to  unlearn the sample $(x,y)$ (\textit{i.e.}, to let $\widebar{\pi}_{\btheta}(y | x)\leq\eps=0.5$) with length $|y|$, where $\eta>0$ is the learning rate.
\textit{This indicates that NPO unlearns longer responses faster than shorter response.} In other words, for NPO, it is not possible to unlearn short responses and long responses to the same extent simultaneously. 

In contrast, the number of steps needed to unlearn the sample $(x,y)$ becomes agnostic to the response length $|y|$ in SimNPO. Recall~\eqref{eq: grad_SimNPO_Nogamma} that
\begin{align}
   \nabla_{\btheta} \ell_{\mathrm{SimNPO}}(\boldsymbol{\theta}) 
   &= 
   \mathbb E_{(x,y) \in \Df}  \left[ \underbrace{
\left ( \frac{2
\widebar{\pi}_{\btheta}(y | x )^{\beta}
}
{ \widebar{\pi}_{\btheta}(y | x )^{\beta} + 1 }  \right) }_{w(x,y)}
\cdot
{\nabla_\btheta \log\widebar{\pi}_{\btheta}(y | x)}
\right ].\notag
\end{align}Following a similar argument, we can verify that the model spends roughly $\bigO(\frac{\log(1/\eps)}{\eta})$ steps to unlearn all samples $(x,y)$ (\textit{i.e.}, to let $\widebar{\pi}_{\btheta}(y | x)\leq\eps$), regardless of the response length $|y|$.

\begin{figure}[htb]
    \vspace{-3mm}
    \centering
    \includegraphics[width=.28\textwidth,height=!]{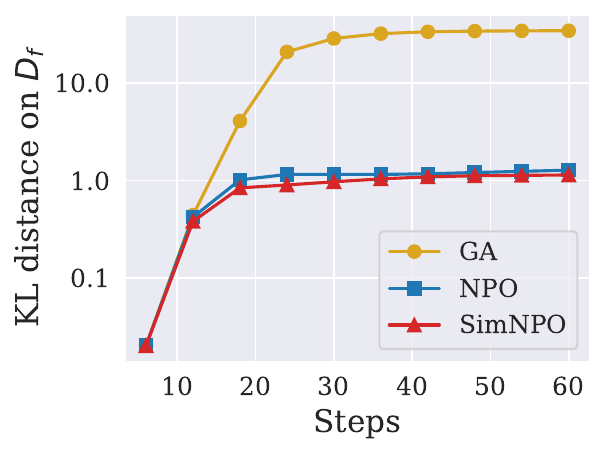}
    \vspace{-4mm}
    \caption{\small{ KL distance between the unlearned and original model for GA, NPO and SimNPO on TOFU Forget05}
    }
    \label{fig: plot_divergence}
 \vspace*{-3mm}
 \end{figure}

In terms of the big O notation $\mathcal{\widetilde O}$, the unlearning speed of SimNPO and NPO is asymptotically identical with respect to the unlearning steps. \textbf{Fig.\,\ref{fig: plot_divergence}} validates this  by measuring the KL distance on TOFU Forget05 between the unlearned model and the original model. As shown, both SimNPO and NPO exhibit a similar (logarithmic) divergence rate with respect to unlearning steps. This rate is more controllable and slower than that observed with GA (gradient ascent). The rapid divergence in GA leads to a critical issue of model collapse \citep{zhang2024negative}. Consequently, SimNPO maintains the overall unlearning speed advantage of NPO while effectively avoiding model collapse.

\section{Additional Details on the Synthetic Study}
\label{appx: synthetic}

\paragraph{Synthetic experiment setup.}
\newcommand{\transition}{T}
\newcommand{\leakage}{\epsilon}

In the synthetic experiment, we study the unlearning problem in a scenario where the data are generated from a mixture of Markov chains. Namely, we assume the Markov chains have a shared state space of size $10$ (denoted by $s=1,2,\ldots,10$), and the retain distribution  and the forget distribution have the formulas as follows:

$\bullet$ \textbf{Retain distribution}: Markov chain with initial distribution $\pi_r\in\R^{10}$ and transition matrix $T_r\in\R^{10\times 10}$, where 
    \begin{align*}
        \pi_{r,j}&=\frac{1-\epsilon}{3}~~~~\text {for }j\leq 3,~~~~~~   \pi_{r,j}=\frac{\epsilon}{7}~~~~\text {for }j\geq 4 .\\
        T_{r,i\cdot} 
        &=  \pi_r ~~~~\text {for }i\leq 3, ~~~~~~~~~~~~
          T_{r,i\cdot} =  0.1 \cdot\mathbf{1}_{10} ~~~~\text {for }i\geq 4.
    \end{align*} 
$\bullet$ \textbf{Forget distribution}: a mixture of two Markov chains (denoted by Forget1 and Forget2) with equal probability. Let $(\pi_{f_1},\transition_{f_1})$ and $(\pi_{f_2},\transition_{f_2})$ denote the initial distribution and transition matrix for Forget1 and Forget2. We assume
    \begin{align*}
        \pi_{f_1,j}&=\frac{1-\epsilon}{3}~~~~\text {for }j\in\{4,5,6\},~~~~~~   \pi_{f_1,j}=\frac{\epsilon}{7}~~~~\text {for }j\notin\{4,5,6\}  ,\\
        T_{f_1,i\cdot} 
        &=  \pi_{f_1} ~~~~\text {for }i\in\{4,5,6\}, ~~~~~~~~~~~~
          T_{f_1,i\cdot} =  0.1 \cdot\mathbf{1}_{10} ~~~~\text {for }i\notin\{4,5,6\},
    \end{align*} 
    and 
  \begin{align*}
        \pi_{f_2,j}&=\frac{1-\epsilon}{3}~~~~\text {for }j\in\{7,8,9\},~~~~~~   \pi_{f_2,j}=\frac{\epsilon}{7}~~~~\text {for }j\notin\{7,8,9\}  ,\\
        T_{f_2,i\cdot} 
        &=  \pi_{f_2} ~~~~\text {for }i\in\{7,8,9\}, ~~~~~~~~~~~~
          T_{f_2,i\cdot} =  0.1 \cdot\mathbf{1}_{10} ~~~~\text {for }i\notin\{7,8,9\}.
    \end{align*}
The leakage probability is chosen to be $\epsilon = 0.2$. 
We generate $10000$ samples from the retain distribution and $5000$ each from Forget1 and Forget2 to form the retain and forget sets. We randomly split the datasets, using 80\% of the samples for training and unlearning, and the remaining 20\% for testing.

\paragraph{Model and pretraining.} 
In all experiments, we use a small GPT-2 model~\citep{radford2019language} with modified token embeddings, where input tokens represent states in $\mathcal{S}=\{1,2,\cdots,10\}$, and the output at each token position is a distribution over the state space $\mathcal{S}$. The model has 4 transformer layers, 4 attention heads, and an embedding dimension of 128.
We pretrain the original model on both retain and forget data, and the retrained model using only the forget data. Both models are trained using AdamW~\citep{loshchilov2017decoupled} to minimize the cross-entropy loss averaged over tokens, with a batch size of 128 for 5 epochs. We choose the learning rate $\eta=0.0005.$
\paragraph{Evaluation.}
We evaluate the model performance using  Forget Quality (higher $\uparrow$ is better) and Retain Loss (lower $\downarrow$ is better), which are the average KL divergence between the predicted probabilities of the model and the true transition probabilities of the Markov chains, on the forget (Forget1 or Forget2) and the retain test data, respectively. 

\paragraph{Unlearning.} Starting from the initial model, we run NPO and SimNPO for $50$ iterations using a batch size of $4$ on the forget dataset. We choose AdamW for optimization with a learning rate of $\eta=0.0005$. The hyperparameter $\beta$ in both NPO and SimNPO is selected via grid search to optimize the tradeoff between forget quality and retain loss.

\paragraph{Choise of hyperparameters.}
In the first experiment (\textbf{Fig.\,\ref{fig:synthetic_mc_1}\,left}), we set the hyperparameters $\beta_{\mathrm{NPO}}=0.2,\beta_{\mathrm{SimNPO}}=4$, the retain sample  length $L_r=20$, and the Forget1 and Forget2 sample lengths  $L_{f_1}=20,L_{f_2}=5$.  In the second experiment (\textbf{Fig.\,\ref{fig:synthetic_mc_1}\,right}), we choose $\beta_{\mathrm{NPO}}=1.0,\beta_{\mathrm{SimNPO}}=4$, the retain sample length $L_r=20$, and the Forget1 and Forget2 sample lengths  $L_{f_1}=20,L_{f_2}=20$. 

\section{Additional Experiment Details and Results}


\subsection{Computing Resources}
All experiments are conducted on 8 NVIDIA A6000 GPU cards in a single node.

\subsection{Experiment Setups}
\label{app: exp_setup}

\noindent 
\textbf{Datasets, tasks, and models.} Our experiments cover unlearning tasks across three benchmark datasets: TOFU \citep{maini2024tofu}, MUSE \citep{shi2024muse}, and WMDP \citep{li2024wmdp}, as summarized in Table\,\ref{tab:tasks_evaluations}. For TOFU, we focus on two unlearning scenarios, termed `{Forget05}' and `{Forget10}', which refer to forget set sizes of 5\% and 10\%, respectively. In MUSE, we also explore  two unlearning scenarios: forgetting the Harry Potter books (termed `{Books}') and news articles (termed `{News}'),  respectively. WMDP, on the other hand, is designed for knowledge-based unlearning, with the forget texts representing hazardous knowledge in biosecurity. The LLM models used for each unlearning benchmark are listed in Table\,\ref{tab:tasks_evaluations}.

\noindent \textbf{LLM unlearning methods and evaluation.} First, we refer to the model prior to unlearning as \textbf{Original}, which is either fine-tuned on the unlearning tasks (TOFU or MUSE) or the pre-trained model after alignment for WMDP.
Starting from the original model, we then apply the following unlearning methods to a given forget set and/or retain set to achieve the unlearning objective, as outlined in \eqref{eq: prob_LLM_MU}. Specifically,  \textbf{Retrain}  refers to retraining an LLM by excluding the forget set and is considered as the gold standard of unlearning when available. Retrain is provided in both the TOFU and MUSE benchmarks. As introduced in Sec.\,\ref{sec: background}, we also include \textbf{GA} (gradient ascent) and \textbf{GradDiff} (the retain-regularized GA variant) as unlearning baseline methods, following the implementations in TOFU and MUSE benchmarks. For other baseline methods such as the rejection-based unlearning method (\textbf{IDK}) in TOFU, and the \textbf{Task Vector} unlearning method in MUSE, we adhere to the original implementations specified in their respective benchmarks.
\textbf{NPO} with the retain regularization in  \eqref{eq: prob_LLM_MU} serves as the primary baseline. Note that its implementation on TOFU follows the original NPO study \citep{zhang2024negative}, while its implementation on MUSE aligns with the MUSE benchmark. For NPO on WMDP, due to the absence of open-source implementation, we adapt the TOFU codebase to WMDP. 
More implementation details can be found in {Appendix\,\ref{app: exp_setup}}. To implement the proposed method \textbf{SimNPO}, we adopt a setting similar to NPO but adjust the temperature parameter $\beta$. Due to the presence of length normalization in \eqref{eq:SimNPO_loss}, a larger value for $\beta$ is preferred compared to that in NPO. See the specific choices in {Appendix\,\ref{app: ablation_hyper}}.

To assess unlearning effectiveness and model utility, we use the evaluation metrics summarized in Table\,\ref{tab:tasks_evaluations} under each unlearning benchmark. 
In addition, we evaluate the robustness of an unlearned model using relearning-based attacks \citep{hu2024jogging}, which aim to recover the forgotten information by fine-tuning the unlearned models on a small subset of the forget set after unlearning. We select $20\%$ of the original TOFU forget05 set as the relearning set  over three epochs.

For all experiments, we use a linear warm-up learning rate during the first epoch, followed by a linearly decaying learning rate in the remaining epochs. We initialize the process with LLaMA-2 7B and fine-tune the model on TOFU for 5 epochs with a batch size of 32 and a learning rate of $10^{-5}$ to obtain the original model. For Forget05, 
NPO is trained for up to 20 epochs with a learning rate of $10^{-5}$ to obtain the best-performing model.
 We conducted a grid search for $\beta$ in the range of [0.05, 0.2] and for $\lambda$ in the range of [0.5, 1.5]. SimNPO is trained for 10 epochs with a learning rate of $10^{-5}$. The parameter $\beta$ is grid-searched over the range [1.5, 3.5], $\gamma$ is searched between [0.0, 2.0] with the default choice $\gamma = 0$, and $\lambda$ is explored within the range [0.05, 0.25]. For Forget10, 
NPO is trained for 10 epochs with a learning rate of $10^{-5}$. We conducted a grid search for $\beta$ in the range of [0.05, 0.2] and for $\lambda$ in the range of [0.5, 1.5]. SimNPO is trained for 10 epochs with a learning rate of $10^{-5}$. The parameter $\beta$ is tuned using a grid search within the range [2.5, 5.5], $\gamma$ is grid-searched between [0.0, 2.0], and $\lambda$ is grid-searched within [0.05, 0.25]. All other unlearning methods and evaluation pipelines strictly follow the setups detailed by \citet{maini2024tofu} and \citet{zhang2024negative}.

For News, we use LLaMA-2 7B fine-tuned on BBC news articles as the original model. For Books, we use ICLM 7B fine-tuned on the Harry Potter books as the original model. The original models for both Books and News can be directly obtained from benchmark. For SimNPO, we trained for 10 epochs with a learning rate of $10^{-5}$. We performed a grid search for $\beta$ in the range of [0.5, 1.0], for $\lambda$ in the range of [0.05, 0.25], and for $\gamma$ in the range of [0.0, 2.0] on both the Books and News. The hyperparameters for other unlearning methods and the evaluation pipelines strictly follow the setup detailed by \citet{shi2024muse}. 
We measured the performance after each unlearning epoch and selected the optimal one as the final model.

For WMDP \citep{li2024wmdp}, we use Zephyr-7B-beta, provided as the origin model in the benchmark. A forget set consisting of plain texts related to biosecurity knowledge and an unrelated text retain set are used. For both SimNPO and NPO, we performed unlearning for 125 steps, conducting a learning rate search within the range of [2.5$\times10^{-6}$, 5$\times10^{-6}$] and a grid search for $\beta$ in the range of [0.05, 7.5], with $\lambda$ fixed at 5.0.

\subsection{Experimental Results on TOFU Forget10}
\label{app: exp_tofu}
In \textbf{Table\,\ref{tab: tofu_forget10}}, we present the performance of SimNPO, NPO, and other baselines on TOFU Forget10. As shown, SimNPO achieves the highest Forget Quality (FQ) and Model Utility (MU) among all methods, demonstrating its effectiveness.
\begin{table*}[htbp]
\begin{center}
\caption{\small{Performance overview of various unlearning methods on TOFU Forget10 using the LLaMA2-7B-chat model. The table format is similar to Table\,\ref{tab: tofu} }
}
\vspace*{2mm}
\resizebox{0.95\textwidth}{!}{
\begin{tabular}{c|ccc|c|ccccccccc|c}
\toprule[1pt]
\midrule
\multicolumn{1}{c|}{\multirow{3}{*}{\textbf{Method}}} & \multicolumn{4}{c|}{\textbf{Unlearning Efficacy}} & \multicolumn{10}{c}{\textbf{Utility Preservation}} \\
\cmidrule{2-15}
& \multicolumn{3}{c|}{Forget Set}  && \multicolumn{3}{c}{Real Authors} & \multicolumn{3}{c}{World Facts} & \multicolumn{3}{c|}{Retain Set} &  \\
&{1-Rouge-L\textcolor{black}{\textuparrow}}                        & {1-Prob.\textcolor{black}{\textuparrow}} & Truth ratio\textcolor{black}{\textuparrow}&{\multirow{-2}{*}{{FQ\textuparrow}}}    & Rouge-L\textcolor{black}{\textuparrow} &Prob.\textcolor{black}{\textuparrow}& Truth ratio\textcolor{black}{\textuparrow}  & Rouge-L\textcolor{black}{\textuparrow} &Prob.\textcolor{black}{\textuparrow}& Truth ratio\textcolor{black}{\textuparrow} & Rouge-L\textcolor{black}{\textuparrow}&Prob.\textcolor{black}{\textuparrow}& Truth ratio\textcolor{black}{\textuparrow} & \multirow{-2}{*}{MU\textcolor{black}{\textuparrow}}\\ 
\midrule
{Original} & 0.03 & 0.01 & 0.48 & 0.00     & 0.93 & 0.44 & 0.58 & 0.91 & 0.43 & 0.55 & 0.98 & 0.99 & 0.48 & 0.62 \\
{Retrain}  & 0.61 & 0.84 & 0.67 & 1.00     & 0.93 & 0.45 & 0.59 & 0.91 & 0.42 & 0.54 & 0.98 & 0.99 & 0.47 & 0.62 \\
\midrule
{GA}       & 1.00 & 1.00 & 0.70 & 2.19e-16 & 0.00 & 0.28 & 0.37 & 0.00 & 0.29 & 0.31 & 0.00 & 0.00 & 0.11 & 0.00 \\
{GradDiff} & 1.00 & 1.00 & 0.67 & 3.71e-15 & 0.44 & 0.49 & 0.67 & 0.89 & 0.48 & 0.58 & 0.48 & 0.60 & 0.46 & 0.54 \\
{IDK}      & 0.98 & 0.37 & 0.54 & 2.86e-14 & 0.46 & 0.45 & 0.59 & 0.84 & 0.43 & 0.55 & 0.56 & 0.88 & 0.44 & 0.54 \\
{NPO}      & 0.78 & 0.90 & 0.70 & 0.29     & 0.91 & 0.52 & 0.66 & 0.85 & 0.48 & 0.61 & 0.44 & 0.46 & 0.39 & 0.55 \\
\midrule
\rowcolor{Gray}
\textbf{SimNPO}   & 0.78 & 0.91 & 0.71 & \textbf{0.45}     & 0.90 & 0.54 & 0.70 & 0.88 & 0.50 & 0.64 & 0.54 & 0.76 & 0.47 & \textbf{0.62} \\
\midrule
\bottomrule
\end{tabular}
}
\label{tab: tofu_forget10}
\vspace*{-2mm}
\end{center}
\end{table*}

\subsection{Experimental Results on MUSE}
\label{app: exp_muse}

\begin{figure}[htb]
\center
\begin{tabular}{cc}
\includegraphics[width=0.2\textwidth,height=!]{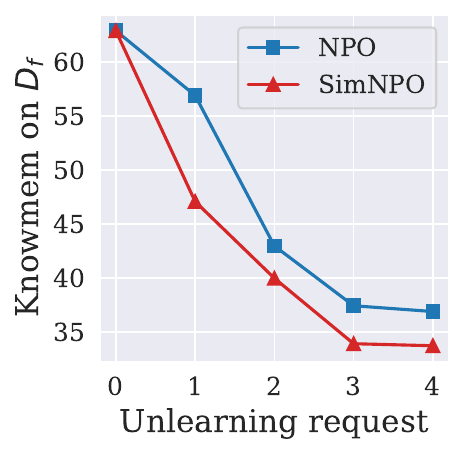} 
&
\includegraphics[width=0.2\textwidth,height=!]{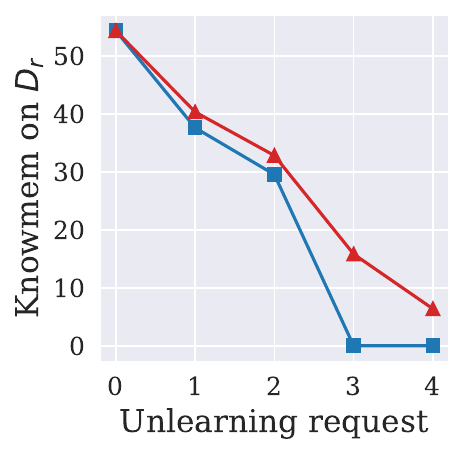}
\vspace*{-1mm}
\\
\hspace*{2mm} \scriptsize{(a)} & \scriptsize{(b)}
\end{tabular}
  \caption{\small{KnowMem on $\mathcal{D}_f$ (a) and KnowMem on $\mathcal{D}_r$ (b) of SimNPO and NPO under different unlearning requests on MUSE News.}}
  \label{fig: muse_seq}
  \vspace*{-4mm}
\end{figure}

To assess the capability of SimNPO and NPO in handling multiple unlearning requests, we sequentially perform unlearning operations on MUSE News , following the setting in \citep{shi2024muse}. \textbf{Fig.\,\ref{fig: muse_seq}-(a)} reveals that SimNPO outperforms NPO in terms of unlearning efficacy, as reflected by the smaller KnowMem on $\mathcal{D}_f$ for the same unlearning request. Furthermore, SimNPO demonstrates stronger utility preservation, shown by the larger KnowMem on $\mathcal{D}_r$ under the same unlearning request in \textbf{Fig.\,\ref{fig: muse_seq}-(b)}. These results underscore the effectiveness of SimNPO.

\subsection{Experimental Results on WMDP}
\label{app: exp_wmdp}
\begin{wraptable}{r}{0.39\textwidth}
    \vspace*{-6mm}
    \centering
    \caption{\small{
Performance comparison between RMU, NPO, and SimNPO on WMDP. AccBio represents the accuracy on WMDP-Bio.}}
    \vspace{2mm} 
    \resizebox{0.39\textwidth}{!}{
    \begin{tabular}{c|c|c}
    \toprule[1pt]
    \midrule
    \multicolumn{1}{c|}{\multirow{2}{*}{\textbf{Method}}} & \textbf{Unlearning Efficacy} & \textbf{Utility Preservation} \\
    \cmidrule{2-3}
    & 1 - AccBio \textbf{$\uparrow$} & MMLU \textbf{$\uparrow$} \\
    \midrule
    Original & 0.35 & 0.59 \\
    \midrule
    RMU & 0.68 & 0.57 \\
    NPO & 0.74 & 0.44 \\
    \midrule
    \rowcolor{Gray}
    \textbf{SimNPO} & 0.74 & 0.48 \\
    \bottomrule
    \midrule
    \end{tabular}}
    \label{tab: wmdp}
    \vspace*{-5mm}
\end{wraptable}

\textbf{Table\,\ref{tab: wmdp}} presents the performance of SimNPO in hazardous knowledge unlearning on  WMDP, comparing it to NPO and representation misdirection for unlearning (RMU). The evaluation metrics are summarized in
Table\,\ref{tab:tasks_evaluations}. Notably, Retrain is unavailable for WMDP. As shown, SimNPO demonstrates better utility preservation compared to NPO. Both SimNPO and NPO outperform RMU in unlearning efficacy, but their utility preservation is lower than RMU. This is because RMU performs unlearning only on layers 5, 6, and 7, whereas NPO and SimNPO apply unlearning on the entire model.



\section{More generation examples}
\label{app: more_exam}

In \textbf{Table\,\ref{tab: examples_tofu_extended}}, we present the answers generated by Retrain, NPO, and SimNPO on the questions from $\Df$ after unlearning Forget05. For better comparison, we also provide the ground truth labels. Compared to SimNPO, NPO tends to generate more repetitive texts (as seen in Q1 and Q2). Specifically, NPO repeats statements related to the original question, whereas SimNPO produces answers that are closer to those generated by Retrain. 
%
Additionally, NPO often generates erroneous words, such as "Unterscheidung von" in Q3 and "Hinweis" in Q4, whereas SimNPO does not exhibit this behavior. Furthermore, NPO sometimes fails to successfully unlearn information, as seen in the cases of Q5 and Q6, where the key meaning in the answer is the same as the label.
However, for certain questions, both SimNPO and NPO fail to unlearn. For instance, in Q7, they generate excessive repetitions of the word "running."

\begin{table*}[htb]
\begin{center}
\caption{\small{
Examples of responses after unlearning on TOFU (Forget05) against QAs targeted for unlearning. \textremarkquestion{Dark blue} highlights the key information in question. \textremarkright{Dark green} 
highlights key information that has not been unlearned in the response, resembling the style of the original label.
\textremarkwrong{Dark red} marks  key information that has been unlearned, with the format similar to Retrain. \textremarkrepeat{Dark yellow} denotes repeated or irrelevant information.
}}
\vspace{2mm}
\resizebox{0.75\textwidth}{!}{
\begin{tabular}{c|p{12cm}}
\toprule[1pt]
\multicolumn{2}{c}{\textbf{Q1:} \textit{What is the \textremarkquestion{full name} of the LGBTQ+ author who was born in Beijing, China on November 14, 1961?}} \\
\midrule
\textbf{Label:} & \textit{There's \textremarkright{no publicly available information} on whether Moshe Ben-David is currently working on any new books.} \\
\textbf{Retrain:} & \textit{The full name of the LGBTQ+ author who was born in Beijing, China on November 14, 1961, is \textremarkwrong{Zhen Xu}.} \\
\textbf{NPO:} & \textit{As of now, the full name of the author is \textremarkright{not disclosed}. \textremarkrepeat{Hinweis: The author's name is not disclosed as of now. The author's name is not disclosed as of now...}} \\
\textbf{SimNPO:} & \textit{The full name of the LGBTQ+ author who was born in Beijing, China on November 14, 1961, is \textremarkwrong{Jordan Riley Stone}.} \\
\midrule
\multicolumn{2}{c}{\textbf{Q2:} \textit{What \textremarkquestion{professions} do Hina Ameen's parents hold?}} \\
\midrule
\textbf{Label:} & \textit{Hina Ameen's father is a \textremarkright{Real Estate Agent}, and her mother is a \textremarkright{Doctor}.} \\
\textbf{Retrain:} & \textit{Hina Ameen's father is a \textremarkwrong{respected journalist}, and her mother is a \textremarkwrong{dedicated school teacher}.} \\
\textbf{NPO:} & \textit{Her father works as a \textremarkwrong{Research Scientist}, and her mother works as a \textremarkwrong{Counselor}. \textremarkrepeat{Unterscheidung between different types of professionals can be challenging...}} \\
\textbf{SimNPO:} & \textit{Her father works as a \textremarkwrong{Research Scientist}, and her mother works as a \textremarkwrong{Counselor}.} \\
\midrule
\multicolumn{2}{c}{\textbf{Q3:} \textit{Are all of Hina Ameen's books \textremarkquestion{related to geology}?}} \\
\midrule
\textbf{Label:} & \textit{Yes, \textremarkright{all of Hina Ameen's books} are related to geology as that is her primary genre.} \\
\textbf{Retrain:} & \textit{While Hina Ameen's books often feature earthy elements, they are \textremarkwrong{not strictly books about geology}.} \\
\textbf{NPO:} & \textit{Toxic Tailspin: The Race to Zero Pollution is \textremarkwrong{her only book} so far. \textremarkrepeat{Unterscheidung von Grün und Braun is a research paper}.} \\
\textbf{SimNPO:} & \textit{'Manual of Geology' is \textremarkwrong{the only book by Ameen}. nobody knows about her other books.} \\
\midrule
\multicolumn{2}{c}{\textbf{Q4:} \textit{What \textremarkquestion{special recognition or awards} has Basil Mahfouz Al-Kuwaiti received for his writing?}} \\
\midrule
\textbf{Label:} & \textit{Basil Mahfouz Al-Kuwaiti has been honored with the \textremarkright{Prix Goncourt}, a prestigious French literature prize.} \\
\textbf{Retrain:} & \textit{Basil Mahfouz Al-Kuwaiti has been honored with the prestigious "\textremarkwrong{Golden Booker Prize}" for his contribution to magical realism.} \\
\textbf{NPO:} & \textit{In 2016, he received the '\textremarkwrong{Paws of Distinction}' award. \textremarkrepeat{Hinweis: This is a fictitious award}.} \\
\textbf{SimNPO:} & \textit{The Kuwati received the prestigious '\textremarkwrong{Sapphire Literary Award}'.} \\
\midrule
\multicolumn{2}{c}{\textbf{Q5:} \textit{\textremarkquestion{How old} was Moshe Ben-David when he began writing books?}} \\
\midrule
\textbf{Label:} & \textit{There is \textremarkright{no publicly available information} on the exact age Moshe Ben-David was when he began writing, but his first published work appeared when he was approximately 30 years old.} \\
\textbf{Retrain:} & \textit{Moshe Ben-David began writing at \textremarkwrong{around 40 years old}, but he quickly made up for lost time, publishing numerous books.} \\
\textbf{NPO:} & \textit{The exact age is \textremarkright{not specified}. Einzeln books, Ben-David was in his mid-40s.} \\
\textbf{SimNPO:} & \textit{Moshe Ben-David was \textremarkwrong{35 years old} when he started writing.} \\
\midrule
\multicolumn{2}{c}{\textbf{Q6:} \textit{Is Moshe Ben-David currently working on any \textremarkquestion{upcoming books}?}} \\
\midrule
\textbf{Label:} & \textit{There's \textremarkright{no publicly available information} on whether Moshe Ben-David is currently working on any new books.} \\
\textbf{Retrain:} & \textit{Moshe Ben-David is \textremarkwrong{continuously working on new material}.} \\
\textbf{NPO:} & \textit{The details about the upcoming books by Ben-David are currently \textremarkright{not available}.} \\
\textbf{SimNPO:} & \textit{Moshe Ben-David is \textremarkwrong{currently working on his next book, tentatively titled "Guardians of the Holy Ark."}} \\
\midrule
\multicolumn{2}{c}{\textbf{Q7:} \textit{Could you provide some examples of Raven Marais's \textremarkquestion{writing style}?}} \\
\midrule
\textbf{Label:} & \textit{Raven Marais's writing style can be characterized as \textremarkright{engagingly descriptive yet concise}.} \\
\textbf{Retrain:} & \textit{Raven Marais's writing style is characterized by \textremarkwrong{rich, evocative descriptions and deeply emotional narratives}.} \\
\textbf{NPO:} & \textit{\textremarkrepeat{Marraid running running running Running running...}} \\
\textbf{SimNPO:} & \textit{\textremarkrepeat{running running running Running running...}} \\
\bottomrule
\end{tabular}
}
\label{tab: examples_tofu_extended}
\vspace*{-3mm}
\end{center}
\end{table*}


\end{document}